\documentclass[runningheads]{llncs}
\PassOptionsToPackage{table,dvipsnames,svgnames,x11names}{xcolor}
 
\usepackage{eccv}

\usepackage{makecell}
\usepackage{xcolor}   
\usepackage{tcolorbox}
\tcbuselibrary{skins}
\usepackage{booktabs}
\usepackage[table]{xcolor}
\usepackage{array}
\usepackage{multirow}
\usepackage{longtable}
\usepackage{graphicx}

\definecolor{deltagray}{gray}{0.90} 
\newcommand{\acc}[2]{\makecell[c]{#1\\[-2pt]{\tiny$\pm$ #2}}}
\newcommand{\accb}[2]{\makecell[c]{\textbf{#1}\\[-2pt]{\tiny$\pm$ #2}}}
\newcommand{\na}{\makecell[c]{---\\[-2pt]{\tiny ---}}}

\usepackage{expl3} %
\usepackage{xparse}

\ExplSyntaxOn
\NewDocumentCommand{\shadegreen}{mmm}{
  \fp_set:Nn \l_tmpa_fp { (#1-#2)/(#3-#2) } 
  \fp_set:Nn \l_tmpa_fp { max(0, min(1, \l_tmpa_fp)) } 
  \fp_set:Nn \l_tmpb_fp { 15 + 70*\l_tmpa_fp } 
  \int_set:Nn \l_tmpa_int { \fp_eval:n { round(\l_tmpb_fp) } } 
  \cellcolor{green!\int_use:N \l_tmpa_int}
}

\ExplSyntaxOff

\usepackage{eccvabbrv}
\usepackage[table]{xcolor}
\usepackage{booktabs}
\usepackage{multirow}
\usepackage{adjustbox}
\usepackage{subcaption}

\usepackage{graphicx}
\usepackage{booktabs}
\usepackage{wrapfig}  

\usepackage{multirow}
\usepackage{adjustbox}
\usepackage{float}
\usepackage{enumitem}

\definecolor{mmbu}{HTML}{A44BA7}
\definecolor{llavamed}{HTML}{FFB540}
\definecolor{llava15}{HTML}{FED475}
\definecolor{medgemma}{HTML}{00b896}
\definecolor{medgemma15}{HTML}{0376ce}
\definecolor{gemma}{HTML}{95E1D3}
\definecolor{medvlmr1}{HTML}{ffbf80}
\definecolor{octomed}{HTML}{e8736d}
\definecolor{qwen2}{HTML}{de40a1}
\definecolor{qwen25}{HTML}{A44BA7}
\definecolor{qwen3}{HTML}{9778BE}
\definecolor{lingshu}{HTML}{C7CEEA}
\definecolor{internvl35}{HTML}{a8dbff}
\definecolor{gpt54}{HTML}{665c5c}
\definecolor{gpt41}{HTML}{918584}
\definecolor{haiku45}{HTML}{E8A882}
\definecolor{sonnet5}{HTML}{D97757}
\definecolor{opus5}{HTML}{B85C38}
\definecolor{opus55}{HTML}{9A4A2E}
\definecolor{sonnet55}{HTML}{C4684A}
\definecolor{gptastra}{HTML}{4A5568}
\definecolor{gpt56}{HTML}{718096}
\definecolor{gpt6sol}{HTML}{536878}
\definecolor{gpt6luna}{HTML}{8497A7}
\definecolor{gpt61sol}{HTML}{334E68}
\definecolor{qwen38}{HTML}{7B68A6}

\newcommand{\totaldatasets}{410}
\newcommand{\uniquemodalities}{11}
\newcommand{\uniquesubmodalities}{35}
\newcommand{\uniquemedicaldomains}{27}
\newcommand{\uniquebodyparts}{95}

\newcommand{\uniquespecimens}{20}

\usepackage[accsupp]{axessibility}  

\usepackage{hyperref}

\usepackage{orcidlink}

\begin{document}

\title{\textcolor{mmbu}{MMBU}: A \textcolor{mmbu}{M}assive \textcolor{mmbu}{M}ulti-modal \textcolor{mmbu}{B}iomedical \textcolor{mmbu}{U}nderstanding Benchmark to Probe the Perception Capabilities of  Vision-Language Models}

\titlerunning{MMBU: Massive Multimodal Biomedical Understanding}


\authorrunning{D'Cunha et al.}

\author{
Ryan D'Cunha\inst{1,*} \and 
Alejandro Lozano\inst{1,}\thanks{Equal contribution.} \and 
Xiaoxiao Sun \inst{1,*}  \and \\
Daniel Vela Jarquin\inst{3} \and
Min Woo Sun\inst{1} \and
Josiah Aklilu\inst{1} \and
James Burgess\inst{1} \and
Yuhui Zhang\inst{1} \and
Mark Endo\inst{1} \and
Ryan Nayebi\inst{1} \and
Akira Nishii\inst{1} \and
Maximilian Rokuss\inst{8} \and
Paola Avila Robayo\inst{1} \and
Jin Ye\inst{4} \and
Ming Hu\inst{4} \and
Zhongying Deng\inst{5} \and
Junjun He\inst{6} \and
Xin Chen\inst{7} \and
Yue Yao\inst{7} \and
Robert Tibshirani\inst{1} \and
Jeffrey J. Nirschl\inst{2} \and
Serena Yeung-Levy\inst{1}
}

\institute{
Stanford University, Stanford, CA, USA 
\and
University of Wisconsin--Madison, Madison, WI, USA
\and
Instituto Tecnológico de Monterrey, Mexico
\and
Monash University, Melbourne, VIC, Australia
\and
University of Cambridge, Cambridge, United Kingdom
\and
Shanghai Jiao Tong University, Shanghai, China
\and
Shandong University, Shandong, China
\and
University of Heidelberg, Germany\\
\texttt{\{rdcunha, lozanoe, xxsun\}@stanford.edu}
}


\maketitle

\vspace{-1.5em} \begin{center}  
\raisebox{-0.2em}{\includegraphics[height=1.1em]{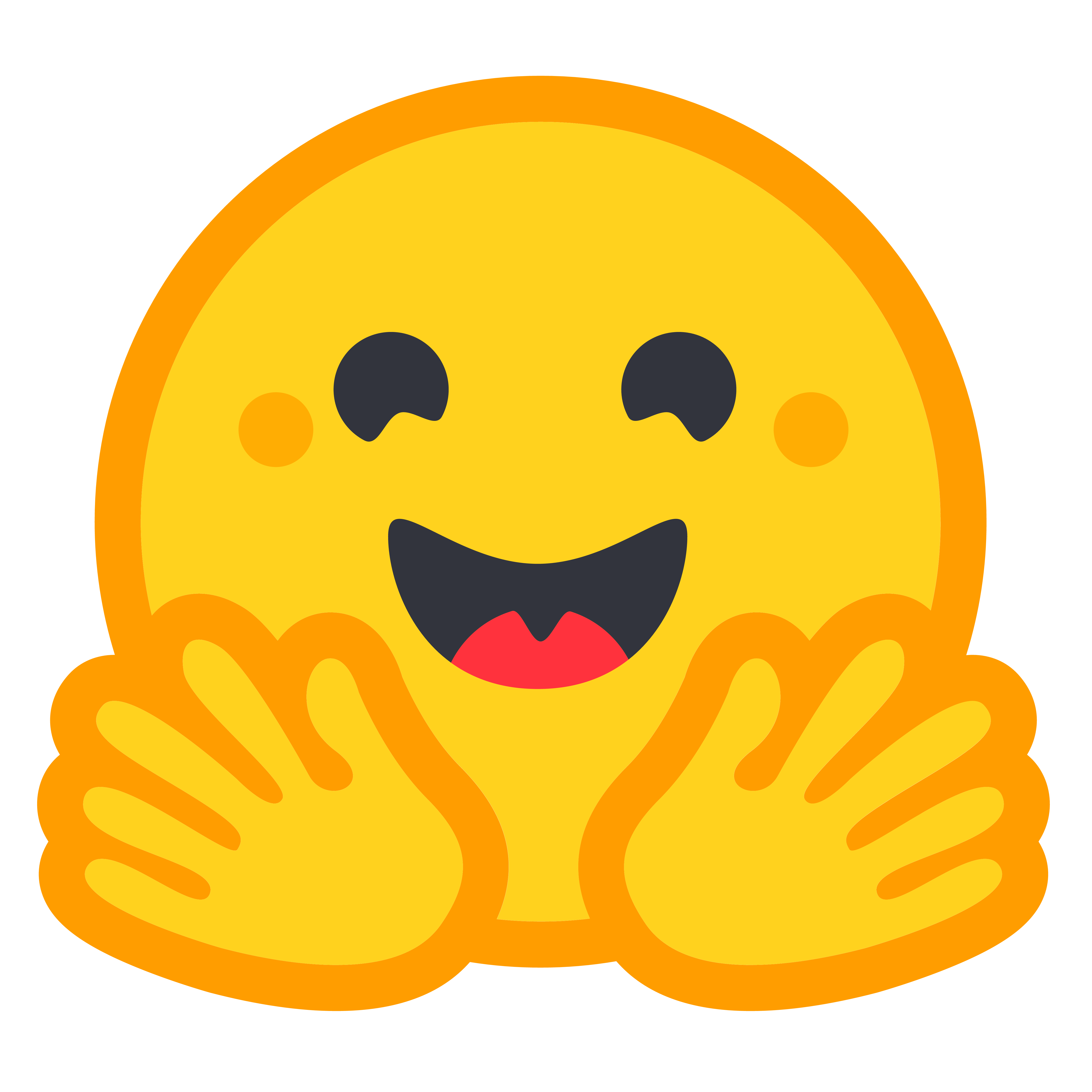}} \ \href{https://huggingface.co/datasets/mmbu/mmbu-public} {\texttt{mmbu/mmbu-public}}
\end{center} 

\begin{figure*}[t]
\begin{center}
	\includegraphics[width=\linewidth]{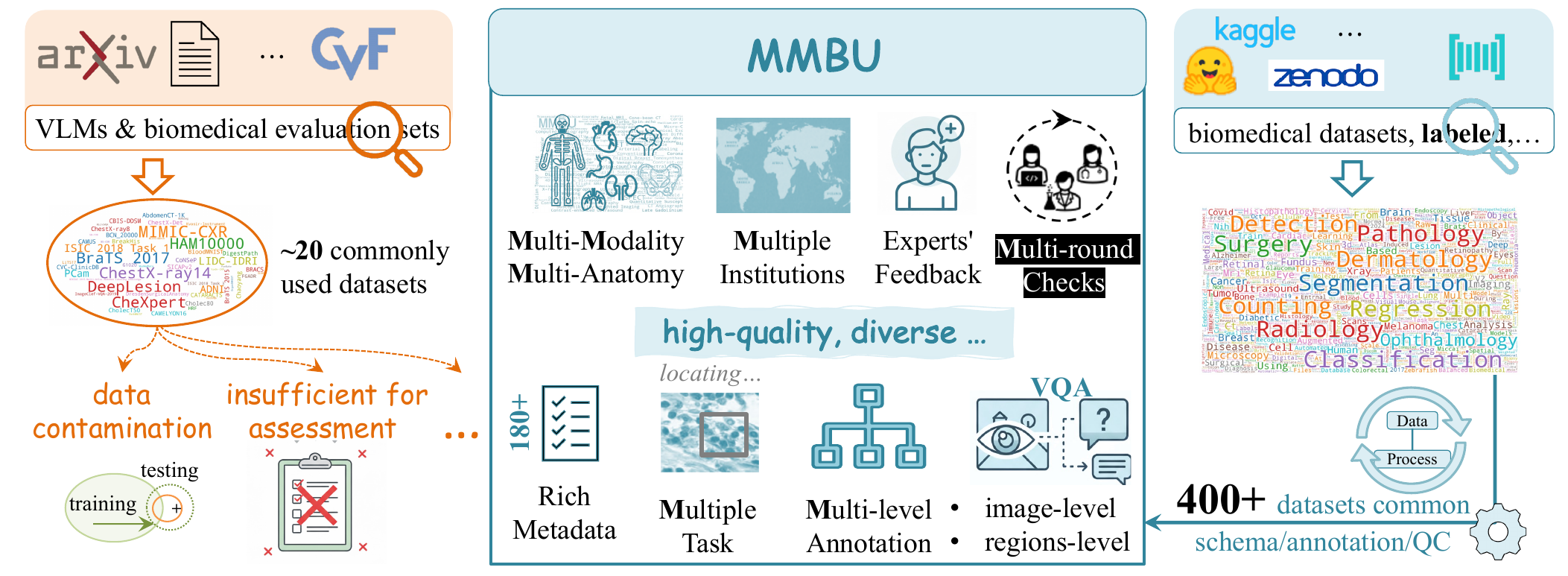}
\end{center}

\caption{\textbf{The data landscape of MMBU.} Current biomedical VLM evaluation relies on roughly 20 commonly used datasets. However, as the training data for large models expands, this evaluation becomes inadequate due to issues such as data pollution and a lack of diversity.  We introduce MMBU to address this issue. }
\label{fig:fig1}
\end{figure*}

\begin{abstract}
Vision and language models (VLMs) hold immense promise to transform biomedical imaging workflows, from detecting lesions in chest X-rays to profiling cellular features in microscopy.  Realizing this potential, however, requires robust and fine-grained visual perception. Models need to correctly interpret subtle features in images, and they must do so across diverse biomedical modalities, scales, and contexts. Nevertheless, current benchmarks remain limited. To address these gaps, we introduce the \textbf{M}assive \textbf{M}ultimodal \textbf{B}iomedical \textbf{U}nderstanding (MMBU) benchmark. It is the largest biomedical vision and language benchmark
to date, covering 35 submodalities with rich structured metadata. It includes both open and closed versions of ungrounded classification, grounded classification, and object detection, enabling systematic evaluation of model performance across biological scales, clinical settings, and imaging modalities. Evaluating 16 open-weight and 5 frontier VLMs in the main comparison, we find that while medical adaptation provides measurable gains for some models, the high accuracy often reported on established benchmarks can
mask deficiencies in visual perception and domain generalization. We further define an open-ended \textbf{MMBU hard set}, split into a released public subset and a held-out private subset, to stress-test frontier models.

\end{abstract}

\section{Introduction}
\label{sec:introduction}

Biomedical vision-language models (VLMs) are increasingly explored for a wide range of biomedical applications~\cite{li2023llava, zhang2023biomedclip}, from clinical decision support~\cite{zhou2025large} to scientific discovery~\cite{maruf2024vlm4bio}. 
To realize this potential, VLMs must first exhibit robust visual perception -- a major failure mode of current VLMs \cite{microvqa} -- as they need to capture both fine- and coarse-grained features and interpret them robustly across heterogeneous patient populations, biomedical modalities, biological scales, and acquisition conditions ~\cite{yang2024limits, varoquaux2022machine}. However, existing evaluations~\cite{pathvqa, slake, vqarad} are often limited to a small set of commonly used datasets and do not provide sufficient metadata to stratify performance to inspect model behavior in detail.  Moreover, the training sets underlying these benchmarks are frequently incorporated into the pretraining or instruction-tuning pipelines of modern biomedical VLMs, leading to substantial distributional overlap between training and evaluation data. As a result, strong performance on established benchmarks may overestimate real-world perceptual capabilities and mask important failure modes.

To rigorously assess visual perception in this setting, benchmarks must go beyond narrow domain-specific evaluation and support systematic analysis across tasks, modalities, specimens, and metadata axes. Motivated by this need, we introduce \textbf{MMBU} (\textbf{M}assive \textbf{M}ulti-modal \textbf{B}iomedical \textbf{U}nderstanding), a large-scale biomedical benchmark designed to evaluate perception capabilities in a broader and more fine-grained way.


Unlike existing single-domain biomedical benchmarks that focus on specific tasks or modalities, such as PathVQA~\cite{pathvqa}, VQA-RAD~\cite{vqarad}, and SLAKE~\cite{slake}, \textbf{MMBU} aims to provide a more reliable evaluation of biomedical VLMs through broader coverage across visual tasks, imaging modalities, biological specimens, and acquisition conditions. 
Such a broader evaluation is necessary, as existing benchmarks cover only a limited portion of the biomedical imaging landscape and recent studies have revealed a gap between benchmark performance and real-world capability. For example, state-of-the-art biomedical VLMs remain highly sensitive to prompt variation and batch effects—systematic shifts arising from differences in scanners, acquisition protocols, or institutions—and often fail to consistently outperform their general-domain counterparts after biomedical adaptation~\cite{gu2025illusion, lozano2024mu, jeong2024limited}. Although these findings are based on relatively small-scale evaluations, they raise important questions about the robustness of current biomedical VLMs and the effectiveness of existing adaptation strategies. By enabling systematic analysis across diverse biomedical modalities, domains, and acquisition conditions, \textbf{MMBU} provides a more comprehensive assessment of biomedical visual perception and helps characterize the limitations of current models.

These findings highlight the need for a benchmark that can systematically evaluate biomedical VLMs across a broader range of visual tasks, domains, topics, and modalities, and thereby more accurately assess when and how domain adaptation improves visual perception. Such an evaluation requires carefully annotated, expert-curated data and a unified task taxonomy that supports comprehensive and comparable analysis. 
To address this need, we assembled a multidisciplinary team of clinicians, bioinformaticians, statisticians, and computer scientists to define a taxonomy of biomedical visual perception tasks and the corresponding domains across biology and medicine. Guided by expert input, we curated data, conducted quality assurance, and iteratively refined a unified benchmark.

\begin{enumerate}

\item \textbf{The \textcolor{mmbu}{MMBU} Benchmark:} Figure \ref{fig:fig1} summarizes the strengths of our contribution. We introduce the largest biomedical VLM benchmark to date (Table~\ref{tab:benchmark_comparison}), covering \uniquemodalities\ modalities (across \uniquesubmodalities\ submodalities) from \uniquespecimens\ specimens and \uniquebodyparts\ unique regions of interest within those specimens, spanning \totaldatasets\ datasets. Each dataset is converted into closed- and open-ended visual question answering (VQA) tasks and manually annotated by a group of experts. MMBU evaluates core perception tasks, including classification and detection, and provides rich, structured annotations for each datapoint, covering image provenance, dataset name, domain, modality, submodality, stain (when applicable), specimen, specimen subregion, topic, original task description, context, institution of acquisition, and URL.

\item \textbf{Systematic evaluation of medically adapted VLMs:} We leverage MMBU to characterize biomedical VLMs (and their base counterparts). Despite steady progress, we find high error rates across all models and tasks, limited but measurable improvement from medical adaptation, substantial weaknesses in spatially grounded tasks such as object detection, and inconsistent generalization beyond commonly used biomedical benchmarks and MMBU.

\end{enumerate}

Given the significant room for improvement, we release MMBU to enable systematic, large-scale evaluation of biomedical VLMs and to support the development of more reliable models for biomedical perception tasks.

\begin{table*}[t]\scriptsize
\centering
\caption{\textbf{Current landscape of biomedical vision–language benchmarks.}  We compare MMBU with existing benchmarks across biomedicine. While prior benchmarks offer valuable but narrow coverage of selected modalities or tasks, MMBU fills a critical gap by unifying a substantially broader range of biomedical domains, submodalities, and metadata sources. $^{*}$If a dataset does not provide the annotation required by a given column, our experts re-annotate the missing fields using the same criteria as MMBU.}

\setlength{\tabcolsep}{0.7pt}{
\begin{tabular}{l c c c c l}
\toprule
\rowcolor{gray!10} \textbf{Benchmark} & \textbf{Submodalities} & \textbf{Size} & \textbf{Topics} & \textbf{Metadata} & \textbf{Source} \\
PathVQA \cite{pathvqa} & 1 & 6K & $\boldsymbol{\times}$ & $\boldsymbol{\times}$ & Textbook, PEIR \\
Cholec80-VQA & 1 & 9K & $\boldsymbol{\times}$ & $\boldsymbol{\times}$ & Cholec80 \\
VQA-RAD \cite{vqarad} & 3 & 3K & $\boldsymbol{\times}$ & $\boldsymbol{\times}$ & MedPix teaching cases \\
SLAKE \cite{slake} & 3 & 2K & $\boldsymbol{\times}$ & $\boldsymbol{\times}$ & MSD, ChestX-ray8, CHAOS \\
RadBench \cite{wright2016radbench} & 6 & 137K & $\boldsymbol{\times}$ & $\boldsymbol{\times}$ & 13 paired image–text datasets \\
MMMU (H \& M) \cite{mmmu} & 6 & 2K & $\boldsymbol{\times}$ & $\boldsymbol{\times}$ & Exams, quizzes, textbooks \\
OmniMedVQA \cite{omnimedvqa} & 12 & 128K & $\boldsymbol{\times}$ & $\boldsymbol{\times}$ & 73 classification datasets \\
GMAI-MMBench \cite{gmaibench}* & 32 & 26K & 321 & 2 & 284 public and hospital datasets \\
\rowcolor{mmbu!10} \textbf{MMBU} & \textbf{\uniquesubmodalities} & \textbf{78K} & \textbf{458} & 13 & \totaldatasets\  curated datasets \\
\bottomrule
\end{tabular}}
\label{tab:benchmark_comparison}
\end{table*}

\section{Related Works}
\label{sec:related_work}

{\bf Biomedical VLMs}
Recent advances in general-purpose VLMs have rapidly expanded their visual reasoning, language grounding, and instruction-following capabilities. These developments have motivated parallel efforts to probe their biomedical strengths \cite{saab2024capabilities,zhou2024evaluating} and to adapt them to medical workflows through instruction tuning or reinforcement learning, leading to models such as LLaVA-Med \cite{li2024llava}, MedGemma \cite{sellergren2025medgemma},  Lingshu \cite{xu2025lingshu}, and OctoMed \cite{ossowski2025octomed}. This accelerating model development underscores the need for rigorous benchmarks that can capture improvements across biomedical domains, modalities, and tasks.

\noindent{\bf Domain-specific biomedical VQA benchmark} Several specialized benchmarks have been developed to evaluate the capabilities of VLMs within specific biomedical imaging modalities—for example, VQA-RAD \cite{vqarad}, U2-Bench \cite{u2bench}, and SLAKE for radiology; PathVQA \cite{pathvqa} for pathology; and MicroBench \cite{lozano2024micro}, and MicroVQA \cite{microvqa}  for microscopy. Although these benchmarks provide detailed, domain-specific assessments and have catalyzed progress within their respective areas, their narrow scope limits their ability to measure how models generalize across the full depth and breadth of biomedical tasks.

\begin{figure}[t]
\begin{center}
	\includegraphics[width=0.9\linewidth]{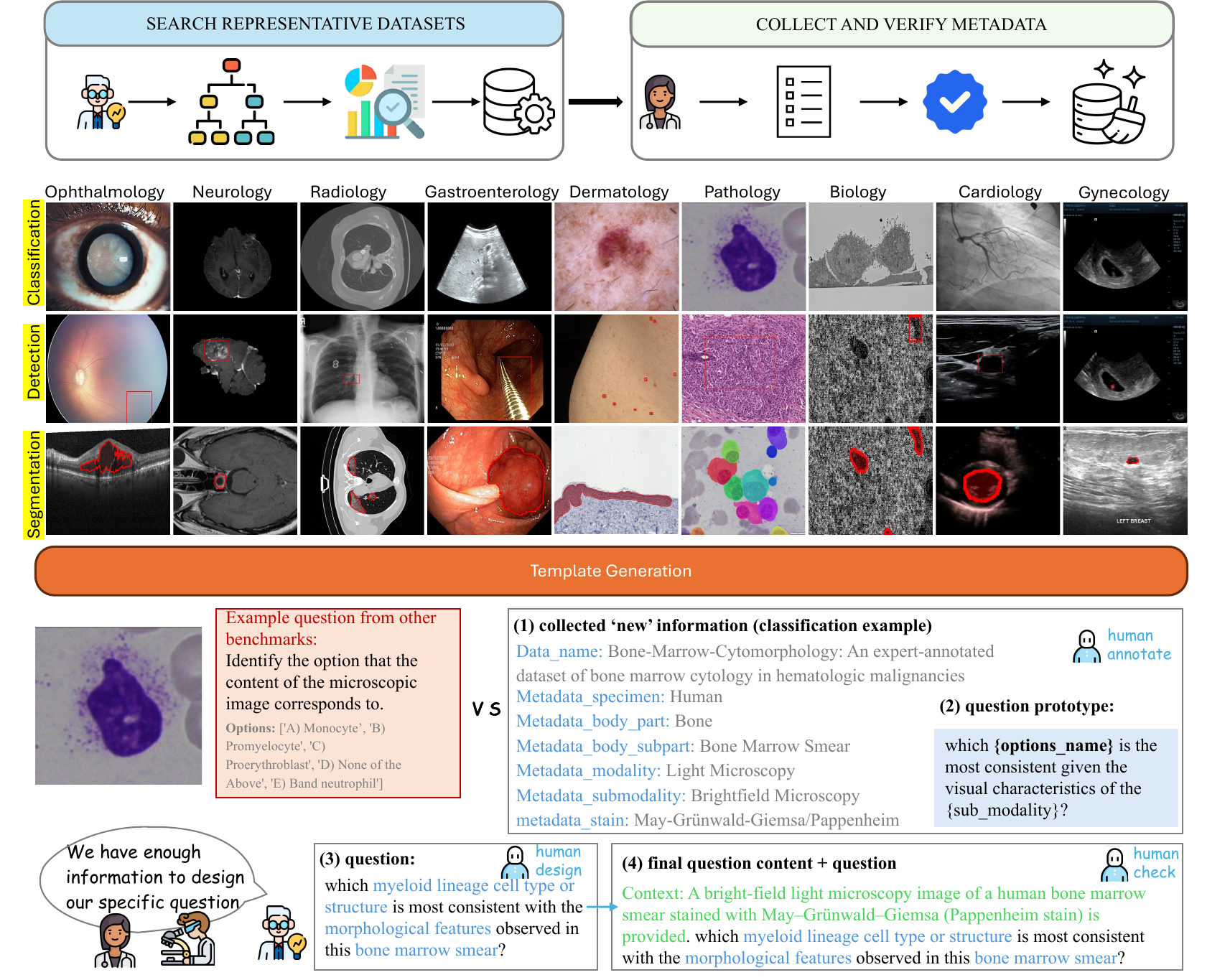}
\end{center}
\vspace{-5mm}
\caption{Multi-task visual examples and metadata-driven question construction in MMBU.
\textbf{TOP: } Data collection metadata extraction and standardization.
\textbf{Middle: }Representative samples from diverse medical domains and modalities across three task types, including classification, detection, and segmentation.
\textbf{Bottom: } Example of question construction in MMBU using newly collected metadata. An example benchmark question is first shown (left), then reformulated through a four-step pipeline (right): (1) collect additional fine-grained metadata from the image and source context (e.g., specimen, body part/subpart, modality, submodality, and stain), (2) define a question prototype conditioned on metadata fields, (3) instantiate a draft domain-specific question targeting clinically meaningful attributes, and (4) produce a finalized question after human design and verification. }
\label{fig:fig-dataset}
\end{figure}

\noindent{\bf Multi-domain biomedical VQA benchmarks}
To support broader evaluation, several recent medical benchmarks span multiple domains, modalities, and tasks. Notable examples include OmniMedVQA \cite{omnimedvqa}, MMMU (Health \& Medicine) \cite{mmmu}, and GMAI-MMBench \cite{gmaibench}. Among these, GMAI-MMBench is one of the most comprehensive, drawing on 284 datasets across  18 clinical tasks. Despite their scale, these benchmarks still have several key limitations: Existing large-scale benchmarks focus primarily on diagnostic imaging (e.g., radiology, pathology, endoscopy) and provide limited coverage of fundamental biological imaging modalities (e.g., fluorescence microscopy, electron microscopy) that reveal underlying cellular and molecular processes driving disease. These benchmarks are largely human-centric, often overlook additional specimens relevant to human health, and are predominantly classification-focused, offering few or no auxiliary tasks. 

Furthermore, they generally lack the contextual information and metadata necessary to stratify performance across meaningful axes; critical details such as provenance, institution, specimen type, topic, and context are frequently absent. For instance, while GMAI-MMBench provides categorization through its lexical tree, MMBU offers more granular and refined annotations, with up to 13 structured attributes per sample encompassing acquisition parameters, task context, and technical factors (e.g., stain). This rich metadata is essential for analyzing model robustness and performance degradation due to batch effects or domain shifts; analyses that are central to MMBU's evaluation objectives but challenging to conduct systematically with existing benchmarks.

\section{Dataset collection methodology}
Our benchmark creation workflow is illustrated in Figure \ref{fig:fig-dataset} and consists of four stages: (1) presenting representative examples spanning classification, detection, and segmentation to demonstrate task and domain diversity in MMBU; (2) collecting fine-grained metadata from images and source context; (3) constructing metadata-conditioned question templates and instantiating draft questions; and (4) performing human-in-the-loop refinement and validation to produce final questions.
This metadata-driven, human-in-the-loop pipeline yields more specific and clinically grounded prompts than directly reusing generic benchmark questions.

\subsection{Data Discovery and Quality Filtering}

We began by designing a multi-granularity dataset taxonomy with input from a multidisciplinary committee of two clinicians, two wet-lab specialists, and two bioinformaticians. The taxonomy was refined through four iterations and covers medical and scientific specialties, image modalities and submodalities, anatomical entities, and biological contexts. Once finalized, we solicited from co-authors representative datasets for each taxonomized item, using the following  inclusion criteria: (i) publicly accessible datasets with clear licensing, (ii) high-quality images, defined as images that are not blurry, where the subject is clearly visible, and without preprocessing artifacts, and (iii) human-generated labels, excluding datasets annotated by LLMs. Additionally, datasets were required to cover core perception tasks (classification, detection, and segmentation). To avoid data contamination, we excluded datasets frequently used to train medical VLMs (e.g., literature-derived datasets). Using these criteria, experts identified 122 relevant datasets in the first round.

In the second round, we developed a discovery agent to search for datasets to increase coverage of underrepresented categories (as identified by our taxonomy) across major public repositories, including Zenodo, Kaggle, Papers with Code, and Hugging Face.  After retrieval, four annotators verified the metadata by cross-checking the original dataset descriptions (and subsequently dropping any dataset not complying with the inclusion criteria); this round identified an additional 1.2K candidate datasets. In the third round, five experts reviewed the aggregated candidate pool and removed datasets with missing annotations, corrupted files, unclear licensing, or preprocessing or formatting errors. This yielded 440 candidate datasets, from which we selected \totaldatasets\ datasets.

\begin{figure*}[t]
\begin{center}
\includegraphics[width=\linewidth]{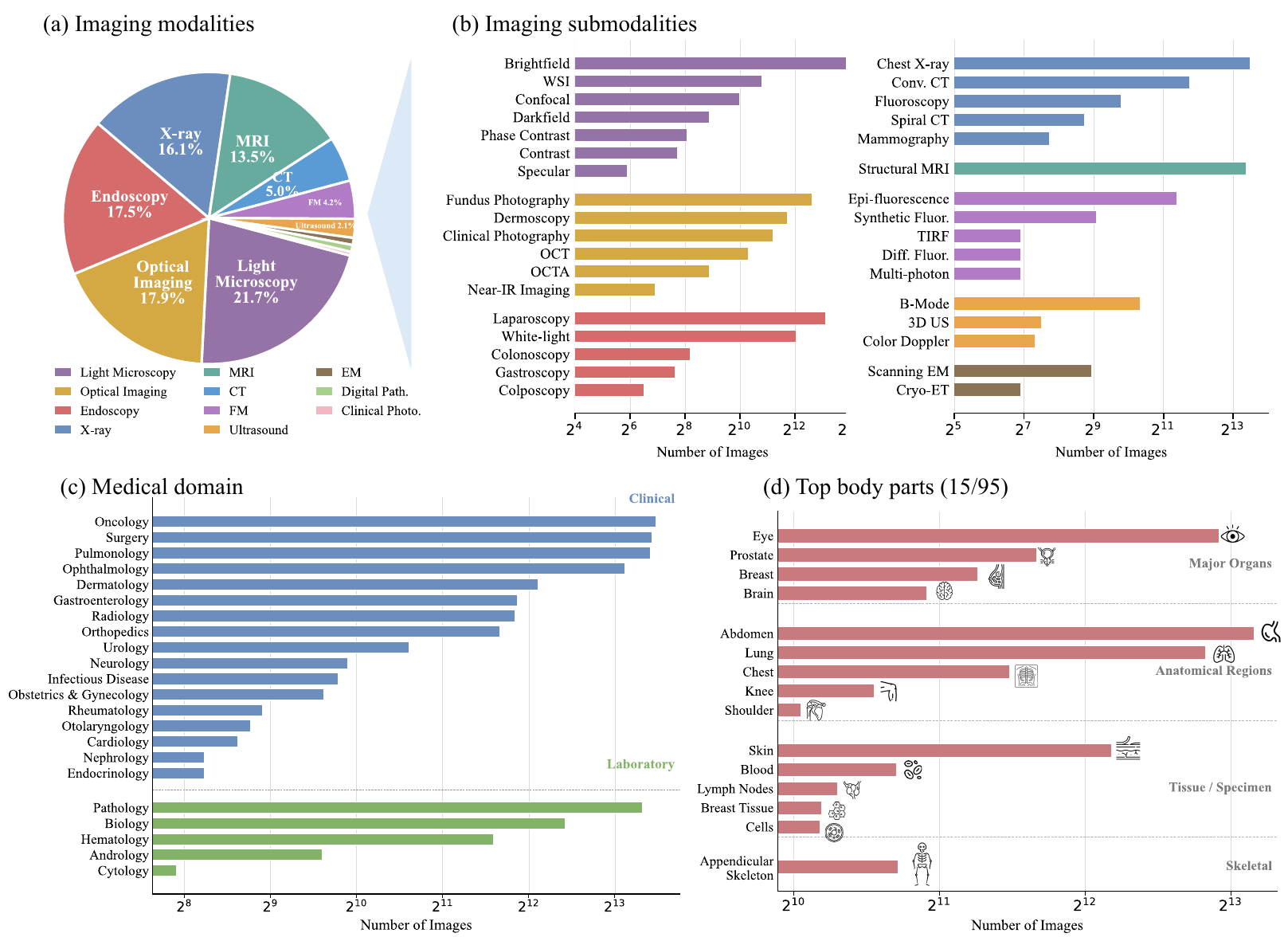}
\vspace{-1.5em}
\end{center}
\caption{Overview of MMBU dataset composition across modalities, submodalities, medical domains, and body parts. (a) Distribution of top-level imaging modalities.
(b) Distribution of imaging submodalities, shown in two panels for readability, with counts on a log2 scale. (c) Distribution of medical domains grouped into clinical and laboratory categories. (d) Top body parts (15 out of 95 total body-part categories), further organized into major organs, anatomical regions, tissue/specimen, and skeletal groups.
Across all panels, bar lengths indicate image counts.}
\label{fig:fig-stats}
\end{figure*}

\subsection{Metadata Collection and Standardization}
The originally collected datasets had varying organizational structures, file formats, and often very limited structured metadata. To address this, we developed a standardized pipeline to harmonize all datasets into a common format. This pipeline handles dataset downloading, integrates annotators to verify and correct extracted metadata, and implements image processing across multiple formats—standardizing all of them to PNG (e.g., TIFF, JPEG, DICOM, NIfTI, SVS, NDPI, MRXS, OME-TIFF).

\subsection{Question Construction, Validation and Refinement}

We provide three versions of templates: two automatic and one manually designed. 
\textbf{No Context:} a standard template shared across all tasks, regardless of dataset. 
\textbf{Modality:} the same template structure, but augmented with the image modality as context in the question. 
\textbf{Full Context:} Annotators used the collected metadata for each dataset to populate a predetermined template. They were asked to include key attributes such as specimen type (e.g., human tissue, human fluid, human smear), imaging modality (e.g., light microscopy, computed tomography, magnetic resonance imaging), stain or contrast when applicable (e.g., hematoxylin and eosin, T1-weighted contrast), preparation method (e.g., smear, section, biopsy), and anatomical site or compartment (e.g., leg, liver, lung). Annotators were further instructed to ensure that all template instructions are grounded in pixel-level visual content, to avoid disease names or clinical outcomes, to avoid any terms suggesting malignancy, and to be clear by not using acronyms. The three context conditions are compared in Section~\ref{sec:supp_context}.

\noindent \textbf{Validation and Refinement.} Lastly, to ensure data quality, all standardized datasets, templates, and metadata are re-evaluated and corrected (if needed) by a team of 3 experts.

\subsection{Dataset Description}
Figure \ref{fig:fig-stats} shows the main MMBU statistics. In summary, MMBU comprises \totaldatasets\ biomedical datasets, including 250 for classification, 84 for segmentation, and 76 for object detection. These datasets span \uniquemodalities\ unique modalities (across \uniquesubmodalities\ submodalities) and cover \uniquemedicaldomains\ medical and scientific domains. Collectively, they include \uniquespecimens\ distinct specimens and \uniquebodyparts\ regions of interest within those specimens.
After standardization, datasets are converted into four categories: ungrounded classification, grounded classification (using segmentation masks), grounded classification (using bounding boxes), and object detection: each having two setups (open and closed), resulting in a total of eight configurations, each with three different templates (24 setups in total). \\

\noindent\textbf{Full-context subset and MMBU hard set.} The full-context evaluation uses approximately 25{,}000 VQA items. From this full-context subset, we curated a representative subset for the MMBU hard-set evaluation track. We then selected a smaller representative public subset containing 3{,}162 open-ended VQA items and retained the remaining 9{,}303 items as \textbf{MMBU-private}, giving 12{,}465 hard-set items in total. \textbf{MMBU-public}, released with answers on Hugging Face\footnote{\url{https://huggingface.co/datasets/mmbu-challenge/mmbu-public}}, supports public benchmarking, while MMBU-private is retained for verified evaluation. The public and private subsets cover the same task families and largely overlapping imaging modalities. The private subset additionally includes sources with licensing restrictions. We report results separately for the eight-task full-context evaluation and the open-ended public/private track. Because the subsets differ in source composition, their score difference should not be interpreted as an isolated measure of difficulty.

\section{Benchmark Experiments}
\subsection{Task Definition}

\textbf{Ungrounded Classification.}
In ungrounded classification, the model predicts a label or answers a question based on the entire image, without any region of interest highlighted or annotated.

\noindent \textbf{Grounded Classification.}
In grounded classification, the model predicts a label for the content within a region of interest, specified by either a segmentation mask or a bounding box.
For both classification tasks, the closed setup requires the model to select the correct answer from a candidate pool, whereas in the open setup, the model must generate the answer directly.

\noindent  \textbf{Object Detection.}
In object detection, the model is tasked with predicting a bounding box given an image and a target class. In the closed setup, the model chooses the correct bounding box from a candidate pool; in the open setup, it must predict the bounding box directly. Scoring is described in Section~\ref{sec:evaluation}.

\subsection{Experimental Setup}

\textbf{Model Selection.} To capture the current VLM landscape, we evaluate 21 autoregressive VLMs in Table~\ref{tab:model_comparison_total} (16 open-weight and 5 frontier API models) and report eight model evaluations under the full-context protocol, together with public/private hard-set results for nine models (Table~\ref{tab:frontier_full_context}), spanning different sizes, instruction-tuning strategies, reasoning capabilities, and medically fine-tuned variants. We compare medically adapted models with the general-purpose counterparts listed below to examine differences associated with biomedical adaptation. Models (color-coded by family throughout the rest of this work) include:
\begin{itemize}[leftmargin=*, itemsep=2pt, topsep=2pt]
    \raggedright
    \item \textcolor{opus55}{\textbf{Opus-5.5}} \cite{anthropic_opus55_2026}, \textcolor{sonnet55}{\textbf{Sonnet-5.5}} \cite{anthropic_sonnet55_2026}, \textcolor{opus5}{\textbf{Opus-5}} \cite{anthropic_opus5_2026}, \textcolor{sonnet5}{\textbf{Sonnet-5} \cite{anthropic_sonnet5_2026}}, and \textcolor{haiku45}{\textbf{Haiku-4.5} \cite{anthropic_haiku45_2025}} from Anthropic.
    \item \textcolor{gptastra}{\textbf{GPT-6-Astra}} \cite{openai_gpt6_astra_2026}, \textcolor{gpt56}{\textbf{GPT-5.6-Sol}} \cite{openai_gpt56_sol_2026}, \textcolor{gpt54}{\textbf{GPT-5.4-mini}} \cite{openai_gpt54mini_2026}, and \textcolor{gpt41}{\textbf{GPT-4.1-mini}} \cite{openai_gpt41mini_2025} from OpenAI. We additionally evaluate \textcolor{gpt6sol}{\textbf{GPT-6-Sol}} \cite{openai_gpt6_sol_2026} and \textcolor{gpt6luna}{\textbf{GPT-6-Luna}} \cite{openai_gpt6_luna_2026} under the same eight-task full-context protocol. Public/private hard-set accuracies for these two models and \textcolor{gpt61sol}{\textbf{GPT-6.1-Sol}} \cite{openai_gpt61_sol_2026} are taken from the MMBU leaderboard~\cite{mmbu_leaderboard_2026}; GPT-6.1-Sol is included only in the hard-set comparison.
    \item \textcolor{lingshu!85}{\textbf{Lingshu-7B}} and \textcolor{lingshu}{\textbf{Lingshu-32B}} \cite{xu2025lingshu}, a family of generalist biomedical VLMs trained in a multi-stage paradigm that progressively infuses medical knowledge, instruction-tuned on over 7.1M multimodal samples. We also evaluate \textcolor{qwen25!85}{\textbf{Qwen2.5-VL-7B-Instruct}} and \textcolor{qwen25}{\textbf{Qwen2.5-VL-32B-Instruct}} \cite{qwen25vltechnicalreport}, the corresponding base models.
    \item \textcolor{octomed}{\textbf{OctoMed-7B}} \cite{ossowski2025octomed}, a recent medical VLM with multimodal reasoning capabilities, built on \textcolor{qwen25!85}{\textbf{Qwen2.5-VL-7B-Instruct}}.
    \item \textcolor{medgemma!60}{\textbf{MedGemma-4B-it}} and \textcolor{medgemma15!70}{\textbf{MedGemma-1.5-4B-it}} \cite{sellergren2025medgemma}, Google's open models for medical text and image comprehension, and the model they were built on, \textcolor{gemma!60}{\textbf{Gemma-3-4B-it}} \cite{gemma3technicalreport}.
    \item \textcolor{llavamed!85}{\textbf{LLaVA-Med-v1.5-7B}} \cite{li2024llava}, one of the first VLMs trained on a large-scale, broad-coverage biomedical figure-caption dataset, and its corresponding base model, \textcolor{llava15!85}{\textbf{LLaVA-1.5-7B}} \cite{llava15}.
    \item One of the strongest open-source general-purpose VLMs, \textcolor{internvl35!85}{\textbf{InternVL3.5-8B}} \cite{internvl35}, which uses a cascaded reinforcement learning approach that enhances reasoning ability via a coarse-to-fine training strategy.
    \item Additional models from the Qwen series, including \textcolor{qwen25!40}{\textbf{Qwen2.5-VL-3B-Instruct}} \cite{qwen25vltechnicalreport}, \textcolor{qwen3!40}{\textbf{Qwen3-VL-4B-Instruct}}, \textcolor{qwen3!60}{\textbf{Qwen3-VL-8B-Instruct}}, \textcolor{qwen3}{\textbf{Qwen3-VL-32B-Instruct}} \cite{qwen3vltechnicalreport}, and the open-weight \textcolor{qwen38}{\textbf{Qwen3.8-27B}} \cite{qwen38modelcard}.
\end{itemize}

This model selection captures variation in size, architecture, reasoning ability, and domain specialization. Open-weight checkpoints are obtained from the respective official Hugging Face repositories and evaluated on NVIDIA L40S and A100 GPUs; proprietary models are accessed through their APIs.

\subsection{Evaluation}
\label{sec:evaluation}

\noindent\textbf{Metrics.}
Table~\ref{tab:model_comparison_total} reports micro-averaged F1 scores with 95\% confidence intervals obtained via 1{,}000-iteration bootstrap resampling over datapoints. Closed-ended answers are scored using the deterministic option parser described below; open-ended classification answers are evaluated for semantic equivalence. For open-ended object detection, correctness requires an Intersection-over-Union (IoU) of at least 0.5 between the predicted and ground-truth bounding boxes. The separate full-context evaluation reports mean task accuracy, as defined below.

\noindent\textbf{Closed-ended VQA answer extraction and scoring}
For closed-ended VQA, we use a fully deterministic extraction and scoring protocol rather than LLM-as-a-judge evaluation \cite{shi2025judging}. 
A hierarchical parser scans model outputs for explicit answer fields, formatting tags, and valid option labels, and then compares the extracted label against the ground-truth option set. 
Responses with no parseable answer or with labels outside the valid option set are counted as incorrect.

\noindent\textbf{Open-ended VQA answer extraction and scoring}
For open-ended classification, exact string matching can reject correct answers expressed using synonymous biomedical terminology.
We therefore use Qwen3-32B as a semantic equivalence judge for free-form textual answers in the main evaluation. We additionally compare judgments from Gemma4-31B and Sonnet-5 to assess sensitivity to judge choice.
The judge receives only the model response and ground-truth answer and is instructed to apply strict biomedical equivalence, including explicit checks for negation, laterality, severity, uncertainty, and diagnostic specificity. The specific judge prompting and information is shown on Hugging Face\footnote{\url{https://huggingface.co/datasets/mmbu-challenge/mmbu-public}}.

\noindent\textbf{Full-context evaluation.} We separately evaluate eight models with structured chain-of-thought prompts and full metadata context on all eight task configurations (approximately 25k items). This comparison includes frontier API models and the open-weight Qwen3.8-27B model. GPT-6-Sol and GPT-6-Luna use the same structured prompts, full metadata context, and Claude Sonnet-5 semantic judging as the other six models. Open-ended responses are evaluated using this common judge. The full-context score is the unweighted mean of the eight task accuracies; public and private hard-set scores are reported separately for their open-ended items. These results use a different prompting and scoring configuration from Table~\ref{tab:model_comparison_total}; differences between the tables should therefore not be attributed solely to model capability. For GPT-6-Sol, GPT-6-Luna, and GPT-6.1-Sol, we additionally report the public/private open-ended hard-set accuracies published on the MMBU leaderboard~\cite{mmbu_leaderboard_2026}. These leaderboard evaluations use a single pass without tools. The eight-task results for GPT-6-Sol and GPT-6-Luna are from our full-context evaluation, not inferred from their hard-set accuracies. GPT-6.1-Sol has no reported full-context or per-task scores in this comparison; those entries are left unreported. Supplementary Section~\ref{sec:supp_protocols} summarizes the distinctions between the reported evaluations.

\section{Results}
\label{sec:results}

\begin{table}[t]
\caption{\textbf{Evaluated VLMs on MMBU tasks.} Micro-averaged F1 scores (with 95\% bootstrap confidence intervals) of 21 representative VLMs and a random baseline on \textit{classification} and \textit{detection} tasks in MMBU. Sub-columns report Closed VQA, Open VQA, and their difference ($\Delta$). \textcolor{Blue}{\textbf{Bold}} indicates the best score per column.}
\label{tab:model_comparison_total}
\centering
{\small
\setlength{\tabcolsep}{3.1pt}
\renewcommand{\arraystretch}{1.35}
\begin{adjustbox}{width=0.98\linewidth}
\begin{tabular}{
l|
cc>{\columncolor{deltagray}}c|
cc>{\columncolor{deltagray}}c|
cc>{\columncolor{deltagray}}c|
cc>{\columncolor{deltagray}}c
}
\toprule
\multirow{2}{*}{\textbf{Model}} &
\multicolumn{3}{c|}{\makecell[c]{\textbf{Ungrounded} \\ \textbf{Classification}}} &
\multicolumn{3}{c|}{\makecell[c]{\textbf{Grounded}\\ \textbf{Classification}\\\textbf{(from Detection)}}} &
\multicolumn{3}{c|}{\makecell[c]{\textbf{Grounded} \\ \textbf{Classification}\\\textbf{(from Segmentation)}}} &
\multicolumn{3}{c}{\textbf{Object Detection}} \\
\cmidrule(lr){2-4}\cmidrule(lr){5-7}\cmidrule(lr){8-10}\cmidrule(lr){11-13}
& Closed & Open & \cellcolor{deltagray}$\Delta$
& Closed & Open & \cellcolor{deltagray}$\Delta$
& Closed & Open & \cellcolor{deltagray}$\Delta$
& Closed & Open & \cellcolor{deltagray}$\Delta$ \\
\midrule

\textcolor{opus5}{\textbf{Opus-5}} &
\textcolor{Blue}{\accb{0.478}{0.017}} &
\textcolor{Blue}{\accb{0.128}{0.004}} &
{\acc{0.351}{0.018}} &
\textcolor{Blue}{\accb{0.325}{0.024}} &
\textcolor{Blue}{\accb{0.247}{0.016}} &
\acc{0.078}{0.028} &
\textcolor{Blue}{\accb{0.418}{0.011}} &
\textcolor{Blue}{\accb{0.086}{0.006}} &
\acc{0.332}{0.013} &
\textcolor{Blue}{\accb{0.203}{0.014}} &
\textcolor{Blue}{\accb{0.143}{0.020}} &
\acc{0.060}{0.024} \\

\textcolor{sonnet5}{\textbf{Sonnet-5}} &
\acc{0.377}{0.011} & \acc{0.084}{0.005} & \acc{0.293}{0.012} &
\acc{0.277}{0.020} & \acc{0.164}{0.017} & \acc{0.114}{0.026} &
\acc{0.405}{0.013} & \acc{0.065}{0.007} & \acc{0.340}{0.015} &
\acc{0.130}{0.009} & \acc{0.075}{0.017} & \acc{0.055}{0.019} \\

\textcolor{haiku45}{\textbf{Haiku-4.5}} &
\acc{0.328}{0.009} & \acc{0.088}{0.005} & \acc{0.241}{0.011} &
\acc{0.228}{0.015} & \acc{0.171}{0.013} & \acc{0.057}{0.020} &
\acc{0.287}{0.007} & \acc{0.065}{0.006} & \acc{0.222}{0.010} &
\acc{0.109}{0.009} & \acc{0.049}{0.013} & \acc{0.060}{0.015} \\

\textcolor{gpt54}{\textbf{GPT-5.4-mini}} &
\acc{0.398}{0.017} & \acc{0.067}{0.005} & \acc{0.331}{0.018} &
\acc{0.287}{0.016} & \acc{0.134}{0.013} &
{\acc{0.153}{0.021}} &
\acc{0.351}{0.011} & \acc{0.058}{0.005} & \acc{0.293}{0.012} &
\acc{0.100}{0.011} & \acc{0.053}{0.014} & \acc{0.047}{0.017} \\

\textcolor{gpt41}{\textbf{GPT-4.1-mini}} &
\acc{0.386}{0.009} & \acc{0.082}{0.004} & \acc{0.305}{0.010} &
\acc{0.287}{0.015} & \acc{0.179}{0.011} & \acc{0.108}{0.019} &
\acc{0.295}{0.007} & \acc{0.065}{0.007} & \acc{0.229}{0.009} &
\acc{0.094}{0.010} & \acc{0.046}{0.013} & \acc{0.048}{0.016} \\

\midrule
\textcolor{gemma!60}{\textbf{Gemma-3-4B}} &
\acc{0.296}{0.010} & \acc{0.042}{0.005} & \acc{0.254}{0.011} &
\acc{0.183}{0.007} & \acc{0.102}{0.013} & \acc{0.081}{0.014} &
\acc{0.323}{0.007} & \acc{0.050}{0.003} & \acc{0.274}{0.008} &
\acc{0.083}{0.007} & \acc{0.025}{0.008} & \acc{0.058}{0.011} \\

\textcolor{medgemma}{\textbf{MedGemma-4B}} &
\acc{0.317}{0.006} & \acc{0.041}{0.003} & \acc{0.276}{0.007} &
\acc{0.240}{0.022} & \acc{0.088}{0.011} & \acc{0.151}{0.025} &
\acc{0.402}{0.017} & \acc{0.043}{0.004} &
{\acc{0.359}{0.018}} &
\acc{0.076}{0.006} & \acc{0.018}{0.008} & \acc{0.058}{0.010} \\

\textcolor{medgemma15}{\textbf{MedGemma-1.5-4B}} &
\acc{0.278}{0.009} & \acc{0.042}{0.003} & \acc{0.236}{0.009} &
\acc{0.204}{0.014} & \acc{0.089}{0.010} & \acc{0.116}{0.017} &
\acc{0.322}{0.006} & \acc{0.044}{0.005} & \acc{0.278}{0.008} &
\acc{0.071}{0.005} & \acc{0.000}{0.000} &
{\acc{0.071}{0.005}} \\

\midrule
\textcolor{internvl35}{\textbf{InternVL3.5-8B}} &
\acc{0.381}{0.007} & \acc{0.110}{0.007} & \acc{0.272}{0.009} &
\acc{0.239}{0.018} & \acc{0.193}{0.021} & \acc{0.046}{0.027} &
\acc{0.313}{0.008} & \acc{0.074}{0.005} & \acc{0.239}{0.009} &
\acc{0.061}{0.005} & \acc{0.019}{0.008} & \acc{0.042}{0.009} \\

\midrule
\textcolor{lingshu}{\textbf{Lingshu-7B}} &
\acc{0.150}{0.010} & \acc{0.047}{0.004} & \acc{0.103}{0.011} &
\acc{0.104}{0.013} & \acc{0.079}{0.012} & \acc{0.024}{0.018} &
\acc{0.170}{0.006} & \acc{0.037}{0.005} & \acc{0.133}{0.008} &
\acc{0.012}{0.007} & \acc{0.001}{0.002} & \acc{0.011}{0.007} \\

\textcolor{lingshu}{\textbf{Lingshu-32B}} &
\acc{0.333}{0.010} & \acc{0.061}{0.005} & \acc{0.273}{0.011} &
\acc{0.213}{0.019} & \acc{0.150}{0.020} & \acc{0.063}{0.027} &
\acc{0.311}{0.012} & \acc{0.053}{0.005} & \acc{0.257}{0.013} &
\acc{0.067}{0.005} & \acc{0.004}{0.004} & \acc{0.063}{0.007} \\

\midrule
\textcolor{qwen25}{\textbf{Qwen2.5-VL-3B}} &
\acc{0.109}{0.007} & \acc{0.025}{0.003} & \acc{0.084}{0.008} &
\acc{0.086}{0.010} & \acc{0.064}{0.008} & \acc{0.022}{0.013} &
\acc{0.184}{0.008} & \acc{0.028}{0.004} & \acc{0.156}{0.009} &
\acc{0.014}{0.003} & \acc{0.017}{0.008} & \acc{-0.002}{0.009} \\

\textcolor{qwen25}{\textbf{Qwen2.5-VL-7B}} &
\acc{0.249}{0.010} & \acc{0.043}{0.004} & \acc{0.206}{0.011} &
\acc{0.184}{0.017} & \acc{0.110}{0.020} & \acc{0.075}{0.026} &
\acc{0.289}{0.008} & \acc{0.046}{0.004} & \acc{0.243}{0.009} &
\acc{0.029}{0.005} & \acc{0.012}{0.006} & \acc{0.017}{0.008} \\

\textcolor{qwen25}{\textbf{Qwen2.5-VL-32B}} &
\acc{0.320}{0.006} & \acc{0.076}{0.008} & \acc{0.243}{0.010} &
\acc{0.213}{0.015} & \acc{0.189}{0.011} & \acc{0.024}{0.019} &
\acc{0.267}{0.007} & \acc{0.064}{0.001} & \acc{0.203}{0.007} &
\acc{0.092}{0.009} & \acc{0.047}{0.012} & \acc{0.045}{0.015} \\

\textcolor{qwen3}{\textbf{Qwen3-VL-4B}} &
\acc{0.293}{0.009} & \acc{0.079}{0.005} & \acc{0.214}{0.010} &
\acc{0.220}{0.022} & \acc{0.161}{0.014} & \acc{0.059}{0.026} &
\acc{0.287}{0.011} & \acc{0.065}{0.006} & \acc{0.222}{0.012} &
\acc{0.040}{0.006} & \acc{0.001}{0.002} & \acc{0.039}{0.006} \\

\textcolor{qwen3}{\textbf{Qwen3-VL-8B}} &
\acc{0.321}{0.011} & \acc{0.088}{0.005} & \acc{0.233}{0.012} &
\acc{0.239}{0.012} & \acc{0.165}{0.021} & \acc{0.074}{0.024} &
\acc{0.291}{0.006} & \acc{0.069}{0.003} & \acc{0.222}{0.007} &
\acc{0.006}{0.003} & \acc{0.001}{0.002} & \acc{0.005}{0.003} \\

\textcolor{qwen3}{\textbf{Qwen3-VL-32B}} &
\acc{0.350}{0.010} & \acc{0.111}{0.004} & \acc{0.240}{0.011} &
\acc{0.241}{0.020} & \acc{0.218}{0.016} & \acc{0.023}{0.026} &
\acc{0.292}{0.009} & \acc{0.082}{0.003} & \acc{0.210}{0.009} &
\acc{0.028}{0.008} & \acc{0.003}{0.004} & \acc{0.024}{0.009} \\

\midrule
\textcolor{llava15}{\textbf{LLaVA-v1.5-7B}} &
\acc{0.254}{0.008} & \acc{0.014}{0.002} & \acc{0.239}{0.008} &
\acc{0.181}{0.016} & \acc{0.037}{0.012} & \acc{0.144}{0.020} &
\acc{0.331}{0.012} & \acc{0.026}{0.003} & \acc{0.305}{0.013} &
\acc{0.031}{0.004} & \acc{0.000}{0.000} & \acc{0.031}{0.004} \\

\textcolor{llavamed}{\textbf{LLaVA-Med-7B}} &
\acc{0.157}{0.012} & \acc{0.018}{0.003} & \acc{0.139}{0.013} &
\acc{0.123}{0.012} & \acc{0.041}{0.011} & \acc{0.082}{0.016} &
\acc{0.250}{0.020} & \acc{0.024}{0.004} & \acc{0.226}{0.021} &
\acc{0.033}{0.007} & \acc{0.002}{0.003} & \acc{0.030}{0.007} \\

\midrule
\textcolor{octomed}{\textbf{OctoMed-7B}} &
\acc{0.351}{0.014} & \acc{0.086}{0.003} & \acc{0.265}{0.015} &
\acc{0.210}{0.013} & \acc{0.154}{0.012} & \acc{0.056}{0.018} &
\acc{0.258}{0.006} & \acc{0.067}{0.004} & \acc{0.191}{0.007} &
\acc{0.033}{0.005} & \acc{0.009}{0.006} & \acc{0.024}{0.008} \\

\textcolor{octomed}{\textbf{OctoMed-2-8B}} &
\acc{0.342}{0.017} & \acc{0.081}{0.005} & \acc{0.261}{0.018} &
\acc{0.221}{0.012} & \acc{0.162}{0.014} & \acc{0.060}{0.019} &
\acc{0.263}{0.005} & \acc{0.058}{0.004} & \acc{0.205}{0.006} &
\acc{0.069}{0.006} & \acc{0.006}{0.005} & \acc{0.063}{0.008} \\

\midrule
\textbf{Random} &
\acc{0.199}{0.004} & \na & \na &
\acc{0.162}{0.003} & \na & \na &
\acc{0.183}{0.004} & \na & \na &
\acc{0.172}{0.004} & \na & \na \\
\bottomrule
\end{tabular}
\end{adjustbox}
}
\end{table}

\begin{figure}[t]
\centering
\includegraphics[width=0.75 \linewidth, trim=4em 0 0 0]{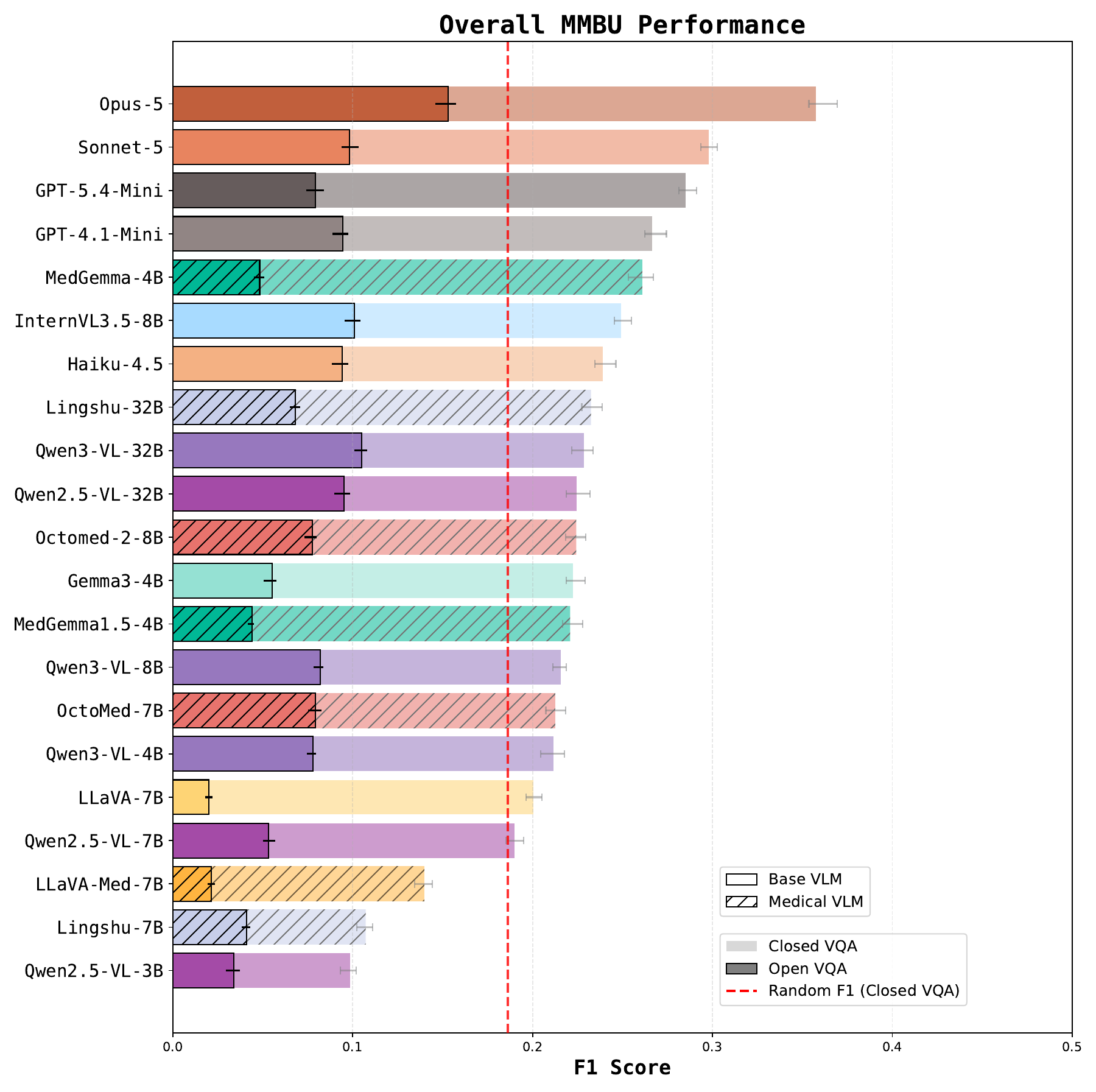}
\caption{\textbf{Aggregate performance on MMBU.} Performance of a representative set of VLMs on all tasks in the benchmark. Solid outlines denote open-format results, while boxes without outlines denote closed-format results. Models are ranked by their closed-format performance. Dashes indicate models adapted to medical data. Similar colors indicate models of the same family.}
\label{fig:overall_by_task}
\end{figure}

\subsection{Overall Performance}
\label{sec:overall_performance}

Table~\ref{tab:model_comparison_total} summarizes the F1 scores of 21 VLMs and a random baseline across four core perception tasks on all of MMBU: ungrounded classification, grounded classification derived from detection annotations, grounded classification derived from segmentation annotations, and object detection. Each task is evaluated under both closed-ended (multiple-choice) and open-ended (free-form) VQA formats; the gap between the two ($\Delta$) quantifies how much performance degrades when the answer space is unconstrained.

Three high-level trends emerge. First, the best closed ungrounded classification F1 is 0.478 (\textcolor{opus5}{Opus-5}), and all closed scores in Table~\ref{tab:model_comparison_total} remain below 0.5. Second, the mean closed-minus-open gap across the three classification families and 21 models is approximately 0.187, using the displayed task-level differences. This gap is consistent with a benefit from answer choices, but also reflects differences between answer recognition, free-form generation, and scoring. Third, object detection remains difficult: only \textcolor{opus5}{Opus-5} exceeds the random closed baseline of 0.172, reaching 0.203, while most open-format detection scores are close to zero.

Larger variants generally improve classification performance within the Qwen2.5-VL and Lingshu families, although gains depend on the task. \textcolor{opus5}{Opus-5} has the highest point estimate in all eight task-format columns of Table~\ref{tab:model_comparison_total}. Among open-weight models, \textcolor{qwen3}{Qwen3-VL-32B} leads open ungrounded classification (0.111). Medical adaptation has mixed effects: \textcolor{llavamed}{LLaVA-Med-7B} scores below \textcolor{llava15}{LLaVA-v1.5-7B} on all three closed classification tasks, but slightly above it on several other task-format combinations. Relative to \textcolor{gemma!60}{Gemma-3-4B}, \textcolor{medgemma15}{MedGemma-1.5-4B} improves detection-derived closed grounded classification from 0.183 to 0.204, while its other closed scores are slightly lower. We examine these task-dependent differences below.

\subsection{Key Observations}
\label{sec:key_observation}

\textbf{High Error Rates Indicate Significant Room for Improvement.} Table~\ref{tab:model_comparison_total} shows that open-ended scores are generally lower than closed-ended scores. Averaging the displayed differences over the 21 models gives gaps of approximately 0.241 for ungrounded classification, 0.077 for detection-derived grounded classification, 0.245 for segmentation-derived grounded classification, and 0.040 for object detection (Table~\ref{tab:delta_f1_scores}). These results indicate substantial remaining difficulty in both answer generation and spatial prediction. The additional task-wise visualization in Figure~\ref{fig:performance_by_task} shows results for the representative model subset included in that figure. Because the open-ended setting is particularly challenging, the following adaptation analyses emphasize closed-ended performance unless specified otherwise.

\begin{figure}[t]
    \centering
    \includegraphics[width=\linewidth]{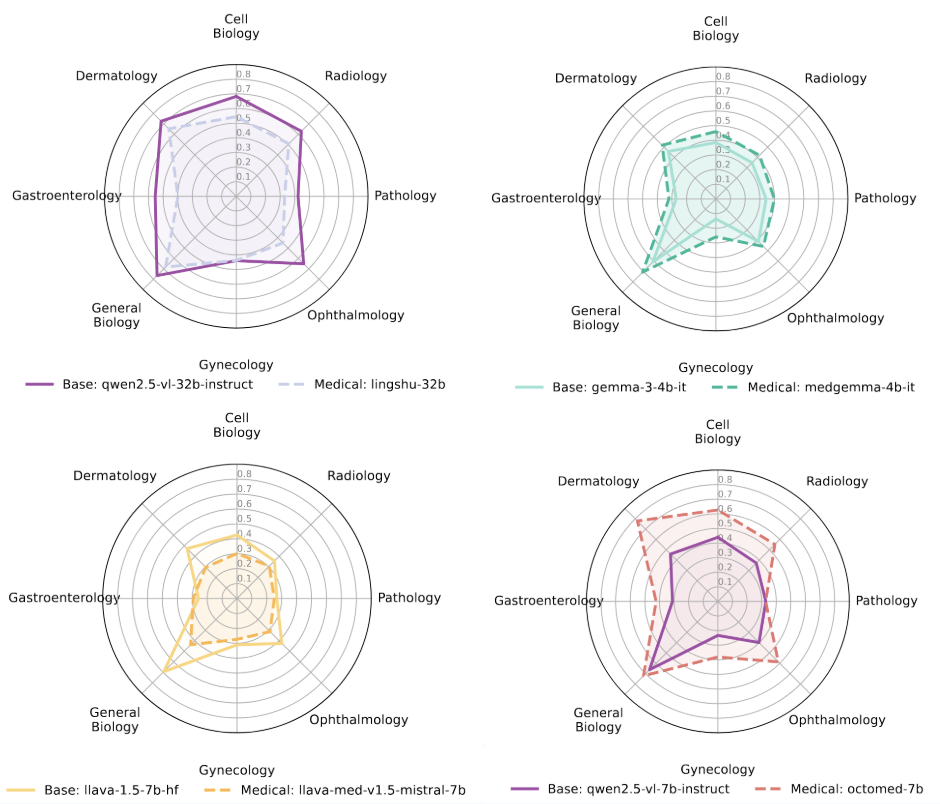}
       \caption{\textbf{Comparison of model performance across aggregated biomedical domains.} 
    The radial plots compare base (solid line) models and their medically adapted (shown by dashes) counterparts across aggregated domains in MMBU.}
    \label{fig:radial_plots_domains}
\end{figure}

\noindent\textbf{Mean Closed Classification Scores Exceed Their Random Baselines.} Equal-weight averages of the displayed scores across the 21 models are approximately 0.308 for ungrounded classification, 0.215 for grounded classification from detection, and 0.300 for grounded classification from segmentation. These means exceed the corresponding random baselines of 0.199, 0.162, and 0.183, although individual models do not always do so. In ungrounded closed classification, \textcolor{opus5}{Opus-5} reaches 0.478 F1 and \textcolor{medgemma}{MedGemma-4B} reaches 0.317, both above the random baseline (0.199). \textcolor{medgemma}{MedGemma-4B} leads open-weight models on segmentation-derived grounded classification (0.402 closed), while \textcolor{opus5}{Opus-5} leads on detection-derived grounded classification (0.325 closed). These results suggest medical specialization may help on some classification axes but does not guarantee superior fine-grained grounded reasoning.

\noindent\textbf{VLMs Exhibit Poor Object Detection Capabilities.}
All open-weight models in Table~\ref{tab:model_comparison_total} score below the closed object detection random baseline of 0.172. Among the frontier API models, \textcolor{opus5}{Opus-5} reaches 0.203. The contrast between grounded classification and detection is consistent with a distinction between recognizing content in a provided region and identifying its location. However, closed detection requires selecting a candidate box, whereas open detection requires generating coordinates, so the two formats also impose different output requirements.

\noindent\textbf{Image vs.\ no-image evaluation.} Following concerns that biomedical VQA items can be answered from language priors alone \cite{asadi2026mirage}, we compare paired image-conditioned and text-only inputs under the legacy ablation protocol. All five models in Table~\ref{tab:image_no_image_main} have higher average closed F1 with images. Nevertheless, closed text-only F1 exceeds the reported random baseline of 0.18 for \textcolor{qwen3}{Qwen3-VL-32B} (0.211), \textcolor{medgemma15}{MedGemma-1.5-4B} (0.201), and \textcolor{internvl35}{InternVL3.5-8B} (0.226). These text-only results indicate that some questions may remain answerable from language priors despite human question construction; the ablation does not isolate the effect of human versus LLM question authorship. The open-ended image benefit varies across models, from 0.003 for OctoMed-7B to 0.101 for Qwen3-VL-32B.

\begin{table}[t]
\centering
\scriptsize
\setlength{\tabcolsep}{4pt}
\caption{\textbf{Image-conditioned vs.\ text-only F1 on MMBU.} Legacy-protocol averages across task families for five open-weight VLMs; $\Delta$F1 = image $-$ no image.}
\label{tab:image_no_image_main}
\resizebox{\linewidth}{!}{
\begin{tabular}{llcc>{\columncolor{gray!15}}c}
\toprule
\textbf{Model} & \textbf{Task set} & \textbf{Image} & \textbf{No image} & \textbf{$\Delta$} \\
\midrule
\multirow{3}{*}{\textbf{OctoMed-7B}}
& Closed VQA & 0.361 & 0.161 & +0.200 \\
& Open VQA   & 0.048 & 0.045 & +0.003 \\
& Overall    & 0.205 & 0.103 & +0.102 \\
\midrule
\multirow{3}{*}{\textbf{Qwen3-VL-32B}}
& Closed VQA & 0.405 & 0.211 & +0.194 \\
& Open VQA   & 0.117 & 0.016 & +0.101 \\
& Overall    & 0.261 & 0.113 & +0.148 \\
\midrule
\multirow{3}{*}{\textbf{MedGemma-4B}}
& Closed VQA & 0.299 & 0.110 & +0.189 \\
& Open VQA   & 0.042 & 0.008 & +0.035 \\
& Overall    & 0.171 & 0.059 & +0.112 \\
\midrule
\multirow{3}{*}{\textbf{MedGemma-1.5-4B}}
& Closed VQA & 0.309 & 0.201 & +0.108 \\
& Open VQA   & 0.032 & 0.009 & +0.023 \\
& Overall    & 0.171 & 0.105 & +0.065 \\
\midrule
\multirow{3}{*}{\textbf{InternVL3.5-8B}}
& Closed VQA & 0.383 & 0.226 & +0.157 \\
& Open VQA   & 0.085 & 0.042 & +0.043 \\
& Overall    & 0.234 & 0.134 & +0.100 \\
\bottomrule
\end{tabular}
}
\end{table}

\begin{figure}[tbp]
    \centering   \includegraphics[width=\linewidth, trim=0 01em 0 0, clip]{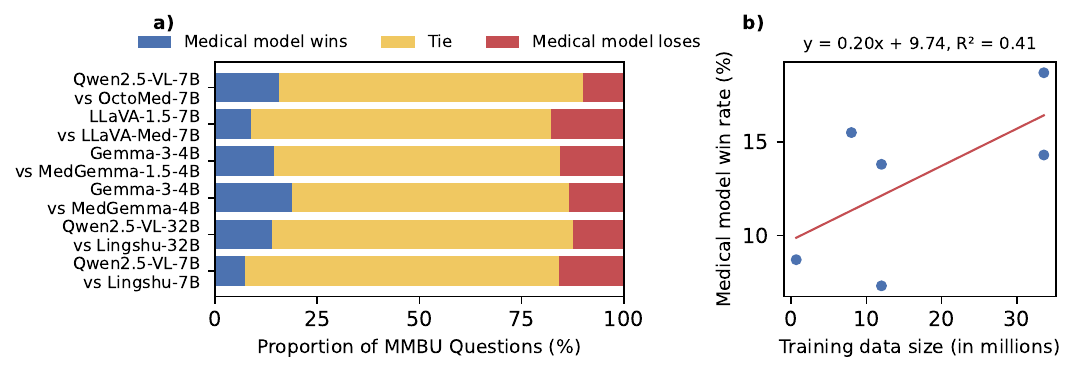}
    \caption{\textbf{Medical models vs. base models.} a) Head‑to‑head comparison of general‑purpose and medical vision–language models on MMBU, showing the proportion of questions where the medical model wins, ties, or loses. b) Relationship between medical model training data size (in millions of examples) and win rate on MMBU.}
    \label{fig:med_v_base}
\end{figure}

\subsection{Full-Context and Hard-Set Evaluation of Frontier and Open-Weight Models}
\label{sec:frontier_full_context}

Table~\ref{tab:frontier_full_context} reports eight-task full-context results for eight models and public/private hard-set accuracies for nine models. Among the eight models with eight-task results (open-ended and closed-ended VQA), \textcolor{opus55}{Opus-5.5} leads (0.453), followed by \textcolor{gptastra}{GPT-6-Astra} (0.425) and \textcolor{sonnet55}{Sonnet-5.5} (0.404). The highest reported eight-task mean remains below 0.46.

On the hard set, \textcolor{gpt61sol}{GPT-6.1-Sol} scores 0.338 on MMBU-public and 0.298 on MMBU-private, placing it third among the nine models reported here on both splits, behind Opus-5.5 and GPT-6-Astra. \textcolor{gpt6sol}{GPT-6-Sol} scores 0.288 and 0.248, while \textcolor{gpt6luna}{GPT-6-Luna} scores 0.254 and 0.218, respectively~\cite{mmbu_leaderboard_2026}. GPT-6.1-Sol exceeds GPT-6-Sol by 5.0 percentage points on each split. All nine models have lower accuracy on the private subset than on the public subset, and private accuracy remains below 0.34.

\begin{table*}[t]
\caption{\textbf{VLMs on full-context MMBU and the open-ended hard set.} Full denotes the unweighted mean accuracy across eight task configurations; Public and Private denote open-ended hard-set accuracies. C and O denote closed and open formats; FG det. and FG seg. denote grounded classification derived from detection and segmentation annotations. For all eight models with full-context results, open responses use Sonnet-5 judging.}
\label{tab:frontier_full_context}
\centering
{\scriptsize
\setlength{\tabcolsep}{2.8pt}
\renewcommand{\arraystretch}{1.2}
\begin{adjustbox}{width=\textwidth}
\begin{tabular}{l|ccc|cc|cc|cc|cc}
\toprule
\textbf{Model} &
\textbf{Full} & \textbf{Public} & \textbf{Private} &
\multicolumn{2}{c|}{\textbf{Cls.}} &
\multicolumn{2}{c|}{\textbf{FG det.}} &
\multicolumn{2}{c|}{\textbf{FG seg.}} &
\multicolumn{2}{c}{\textbf{Det. bbox}} \\
& & & & C & O & C & O & C & O & C & O \\
\midrule
\textcolor{opus55}{\textbf{Opus-5.5}} &
\textbf{0.453} & \textbf{0.370} & \textbf{0.336} &
0.580 & 0.322 & 0.468 & 0.326 & 0.557 & 0.517 & 0.612 & 0.243 \\
\textcolor{gptastra}{\textbf{GPT-6-Astra}} &
0.425 & 0.348 & 0.304 &
0.469 & 0.283 & 0.437 & 0.284 & 0.540 & 0.495 & 0.598 & 0.296 \\
\textcolor{sonnet55}{\textbf{Sonnet-5.5}} &
0.404 & 0.318 & 0.274 &
0.553 & 0.262 & 0.416 & 0.243 & 0.524 & 0.455 & 0.567 & 0.215 \\
\textcolor{gpt56}{\textbf{GPT-5.6-Sol}} &
0.365 & 0.261 & 0.238 &
0.538 & 0.238 & 0.409 & 0.265 & 0.507 & 0.294 & 0.483 & 0.186 \\
\textcolor{opus5}{\textbf{Opus-5}} &
0.361 & 0.279 & 0.270 &
0.533 & 0.272 & 0.342 & 0.269 & 0.496 & 0.345 & 0.488 & 0.140 \\
\textcolor{qwen38}{\textbf{Qwen3.8-27B}} &
0.272 & 0.197 & 0.174 &
0.468 & 0.169 & 0.325 & 0.141 & 0.483 & 0.334 & 0.243 & 0.013 \\
\midrule
\textcolor{gpt61sol}{\textbf{GPT-6.1-Sol}} &
--- & 0.338 & 0.298 &
--- & --- & --- & --- & --- & --- & --- & --- \\
\textcolor{gpt6sol}{\textbf{GPT-6-Sol}} &
0.396 & 0.288 & 0.248 &
0.534 & 0.228 & 0.405 & 0.247 & 0.514 & 0.423 & 0.591 & 0.229 \\
\textcolor{gpt6luna}{\textbf{GPT-6-Luna}} &
0.343 & 0.254 & 0.218 &
0.480 & 0.201 & 0.355 & 0.195 & 0.470 & 0.393 & 0.468 & 0.182 \\
\bottomrule
\end{tabular}
\end{adjustbox}
}
\end{table*}

\subsection{Additional Findings and Discussion}
\label{sec:findings_discussion}

Beyond the aggregate results in Table~\ref{tab:model_comparison_total}, we investigate two questions central to the current biomedical VLM landscape: (i) does medical adaptation consistently improve performance over general-purpose base models, and (ii) do gains observed on established benchmarks transfer to the broader evaluation provided by MMBU? 

\noindent\textbf{Medical Models Show Advantages in Selected Submodalities.} Stratifying performance by metadata identifies submodalities in which medically adapted models have an advantage. For example, when evaluating performance by submodality, medical models outperform non-medical models 62.28\% of the time (Supplementary Materials)
with an average F1 lead of only 0.08. We see the best performance from medical models on Near-Infrared Imaging, Phase Contrast Microscopy, and Specular Microscopy.

\noindent\textbf{Medical adaptation yields limited but measurable gains.} 
To assess adaptation within model families, we compare medically adapted models with their listed general-purpose counterparts. 
Figure \ref{fig:med_v_base} (a) compares the closed classification performance of base models and their medically adapted counterparts. 
Ties account for most comparisons between medically adapted models and their general-purpose counterparts. Notably, only about half of the adapted models outperform their base counterparts, with the strongest gains observed in \textcolor{medgemma}{MedGemma-4B} (18.7\% wins / 13.4\% losses), \textcolor{octomed}{OctoMed-7B} (15.5\% wins / 9.9\% losses), and \textcolor{lingshu}{Lingshu-32B} (13.8\% wins / 12.4\% losses). 
Figure \ref{fig:radial_plots_domains} further stratifies performance by a reduced set of domains of interest. Indeed, only \textcolor{octomed}{OctoMed-7B} and \textcolor{medgemma}{MedGemma-4B} consistently exhibit improvements with respect to their base counterparts.
Across the plotted model pairs, training data size is positively associated with win rate (Figure~\ref{fig:med_v_base}(b); $R^2 = 0.41$). This observational comparison does not isolate the effect of training data size from architecture, model size, or training procedure.

These results highlight that medical adaptation is feasible, although substantial gains remain limited when evaluated broadly, partially aligning with recent findings showing that adapted models often underperform their base counterparts more frequently than they outperform them \cite{jeong2024limited}.


\begin{figure*}[t]
    \centering
\includegraphics[width=\linewidth, trim=0 0 4em 0]{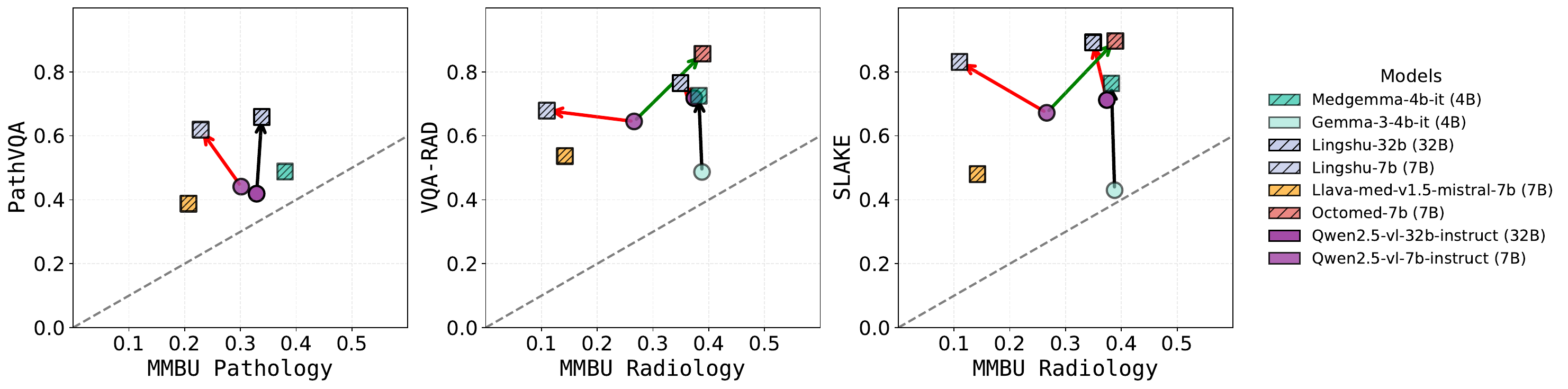}
   \caption{\textbf{Performance comparison across MMBU subsets (x-axis) and legacy benchmarks (y-axis).} Arrows indicate changes after medical adaptation. The comparison examines whether changes on legacy benchmarks (PathVQA, VQA-RAD, SLAKE) also occur on the corresponding MMBU subsets. Improvement on a legacy benchmark does not imply improvement on MMBU.}
   
    \label{fig:corr_trad_vs_mmbu}
\end{figure*}

\noindent\textbf{Cross-Dataset Generalization of Medically Adapted VLMs.}
Figure \ref{fig:corr_trad_vs_mmbu} compares base and adapted VLMs on legacy benchmarks (PathVQA, VQA-RAD, SLAKE) and the corresponding MMBU subsets. While all adapted models improve on legacy benchmarks, gains do not consistently transfer to MMBU: \textcolor{octomed}{OctoMed} improves on both; \textcolor{medgemma}{MedGemma} improves on legacy benchmarks but remains flat on MMBU; and \textcolor{lingshu}{Lingshu} models improve on legacy benchmarks but degrade on MMBU. These results show that gains on the evaluated legacy benchmarks do not uniformly transfer to the corresponding MMBU subsets. Exposure to related training data is one possible explanation, since widely used benchmark training sets appear in medical fine-tuning~\cite{li2024llava} or pretraining~\cite{xu2025lingshu}; this comparison does not isolate that explanation from other differences between benchmarks.


\section{Conclusion}
\label{sec:conclusion}
MMBU exposes pervasive weaknesses in biomedical VLMs’ perceptual abilities: low ungrounded and grounded classification accuracy, poor detection, and inconsistent cross-dataset generalization, despite increasing model size or adaptation. While no benchmark can replace expert evaluation, MMBU provides a representative proxy for measuring progress. By consolidating diverse perceptual tasks and rich metadata, it highlights clear directions for future work: stronger spatial modeling, improved perception, and adaptation methods that generalize beyond currently established datasets.

\bibliographystyle{splncs04}
\bibliography{main}

\newpage

\section*{Supplementary Material}
The supplement presents examples of the four open-ended task families, summarizes dataset composition and annotator expertise, identifies the evaluated model families, and distinguishes the evaluation protocols. We then report task, modality, and domain breakdowns; compare question-context and image conditions; and examine embedding visualizations and matched-image question formulations.

\section{Acknowledgements}
We would like to acknowledge Stanford Data Science, the NVIDIA Academic Grant Program, Anthropic AI for Science, and AWS for providing computing resources for this project. This research was supported by NIH grants (NIH \# P30AG066515 to JJN), a Hoffman-Yee Research Grant to SYL, the Arc Institute Graduate Fellowship to AL, the Stanford Data Science Scholars program to MW and the Quad Fellowship to JB. SYL is a Chan Zuckerberg Biohub – San Francisco Investigator.

\section{Examples of Open-Ended MMBU Tasks}
\label{sec:supp_task_examples}
\begin{figure*}[p]
    \centering
    \includegraphics[width=\textwidth,height=0.82\textheight,keepaspectratio]{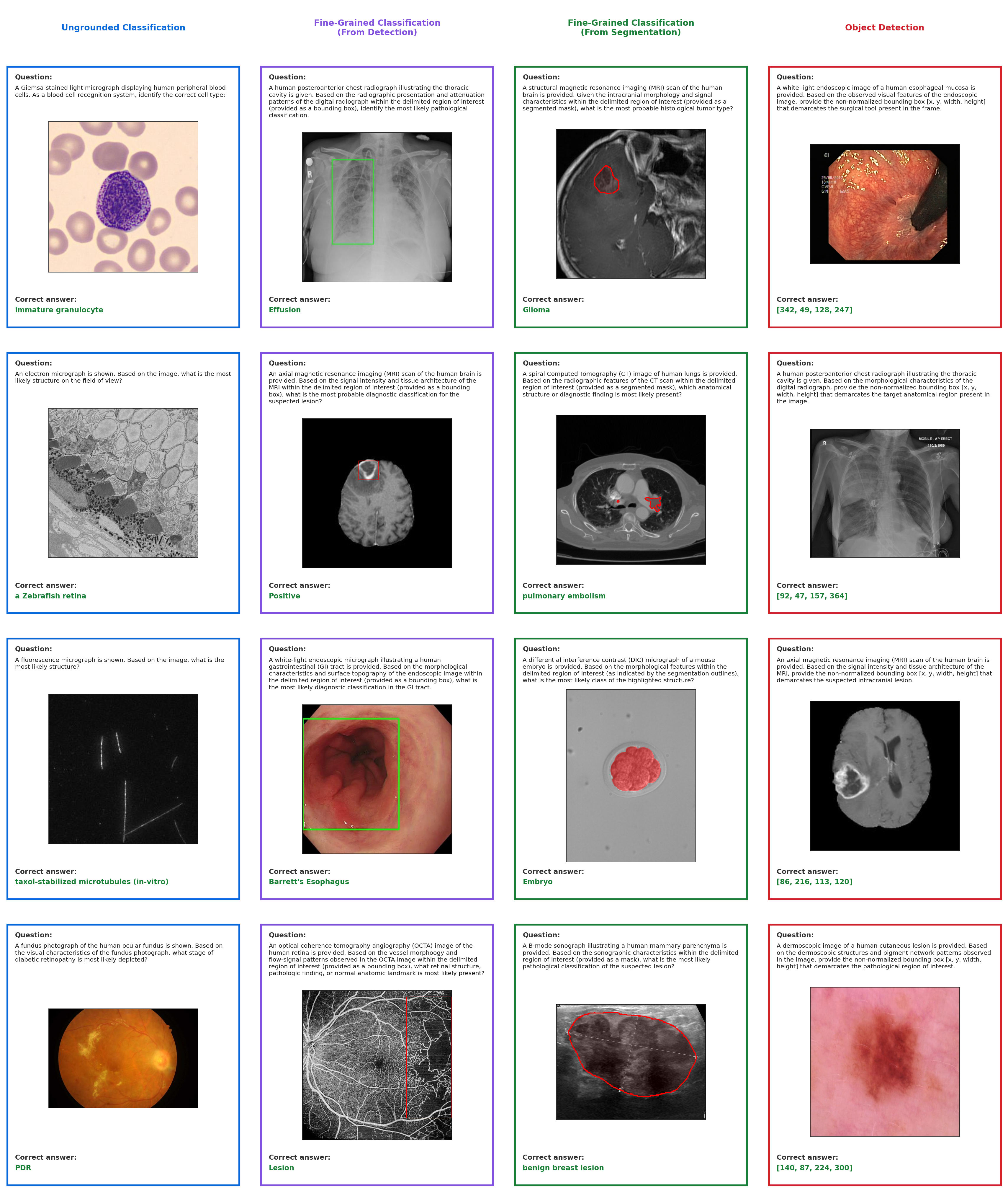}
    \caption{\textbf{Examples of open-ended MMBU tasks.} From left to right: ungrounded classification, grounded classification from detection annotations, grounded classification from segmentation annotations, and object detection. The two middle columns are labeled ``Fine-Grained Classification'' in the figure and correspond to the grounded classification tasks defined in the main text. Boxes and mask overlays identify the supplied regions for grounded classification. Detection answers use non-normalized bounding-box coordinates in $[x,y,\mathrm{width},\mathrm{height}]$ format for the associated input image. Green answer text denotes the reference answer.}
    \label{fig:open_vqa_task_examples}
\end{figure*}

Figure~\ref{fig:open_vqa_task_examples} presents 16 examples spanning the four MMBU task families, with four examples per family. Each example pairs an image and question with its reference answer. The examples illustrate the task inputs and expected answer formats rather than model predictions.

\clearpage

\section{Additional Dataset Statistics}
\label{a:dataset-stats}

\begin{figure}
    \centering
    \includegraphics[width=0.95\linewidth]{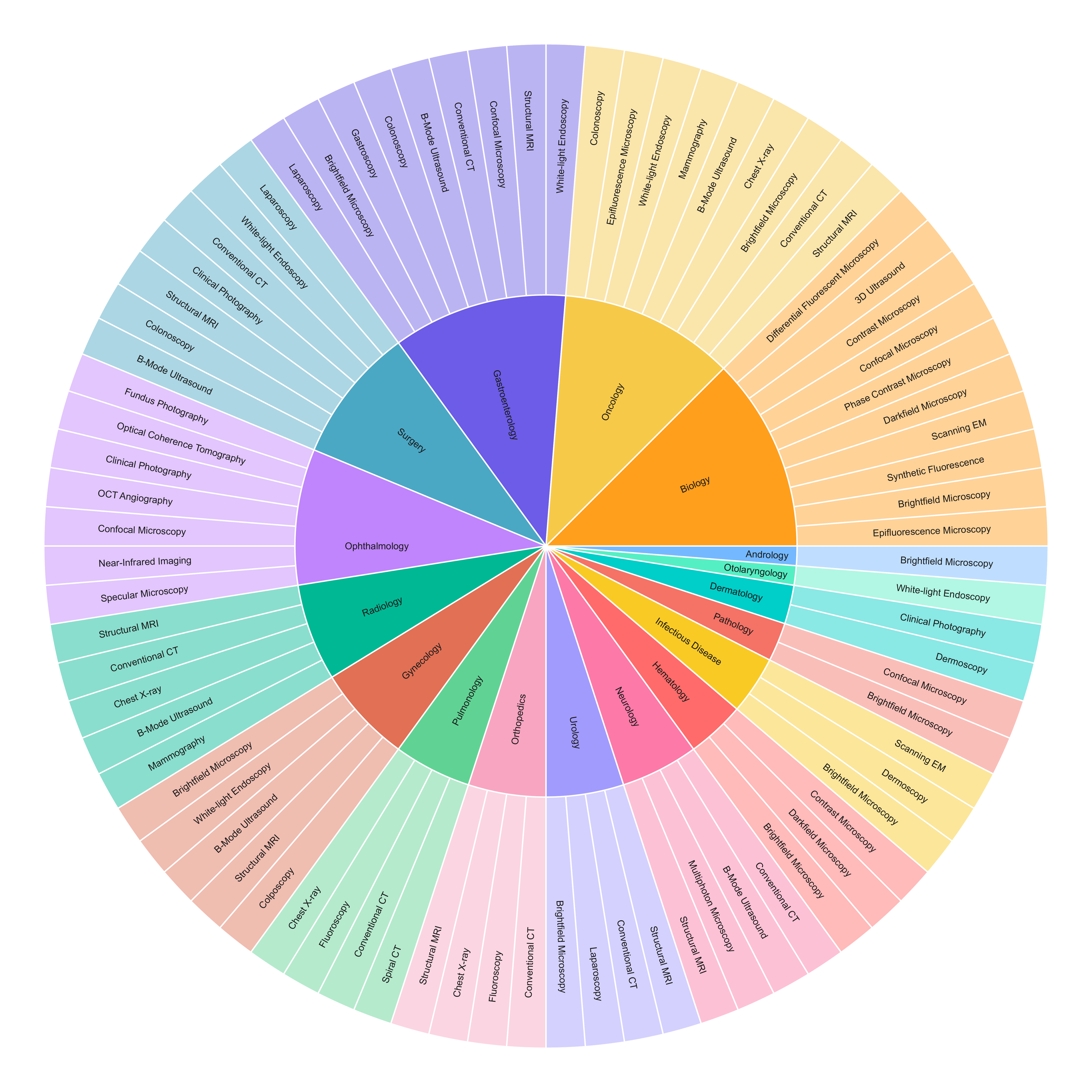}
    \caption{\textbf{Domain and submodality coverage in MMBU.} The inner ring identifies biomedical domains; the outer ring lists the imaging submodalities represented within each displayed domain.}

    \label{fig:modality_submodality}
\end{figure}

\begin{table}[H]
\centering
\caption{MMBU  Summary Statistics}
\begin{tabular}{lc}
\toprule
\textbf{Aspect} & \textbf{Count} \\
\midrule
Collected Datasets     & \totaldatasets  \\
Unique Modalities         & \uniquemodalities \\
Unique Submodalities      & \uniquesubmodalities\\
Unique Domains  & \uniquemedicaldomains \\
Unique Regions        & \uniquebodyparts \\
Unique Annotators               & 8  \\
\bottomrule
\end{tabular}
\label{tab:medalign_statistics}
\end{table}

Table~\ref{tab:medalign_statistics} summarizes the benchmark's source datasets, modality coverage, domains, regions, and annotator count. Figure~\ref{fig:modality_submodality} shows how the displayed biomedical domains map to imaging submodalities.

The annotator pool (Table~\ref{tab:expert_annotators}) spans both clinical and technical domains, covering clinical pathology, surgery, biomedical engineering, and biomedical informatics. The table describes the expertise available during curation; annotation consistency also depends on the review and adjudication procedure.

\begin{table}[h!]
\centering
\caption{\textbf{Annotator expertise and domain background.} Summary of the annotators involved in MMBU curation and verification, including years of experience and primary domain expertise.}
\label{tab:expert_annotators}
\begin{tabular}{|c|c|c|}
\hline
Annotator ID & Years of Experience & Domain \\
\hline
A001 & 10 & Clinical Pathology  \\
A002 & 12 & Surgery \\
A003 & 4 &  Biomedical Engineering  \\
A004 & 5 & Biomedical Informatics \\
A005 & 3 & Biomedical Engineering \\
A006 & 2 & Biomedical Informatics \\
A007 & 2 & Biomedical Informatics \\
A008 & 1 & Biomedical Informatics \\
\hline
\end{tabular}
\end{table}

\subsection{Public and Private Split Visualization}
\label{sec:supp_public_private}
\begin{figure*}[t]
    \centering
    \includegraphics[width=\textwidth]{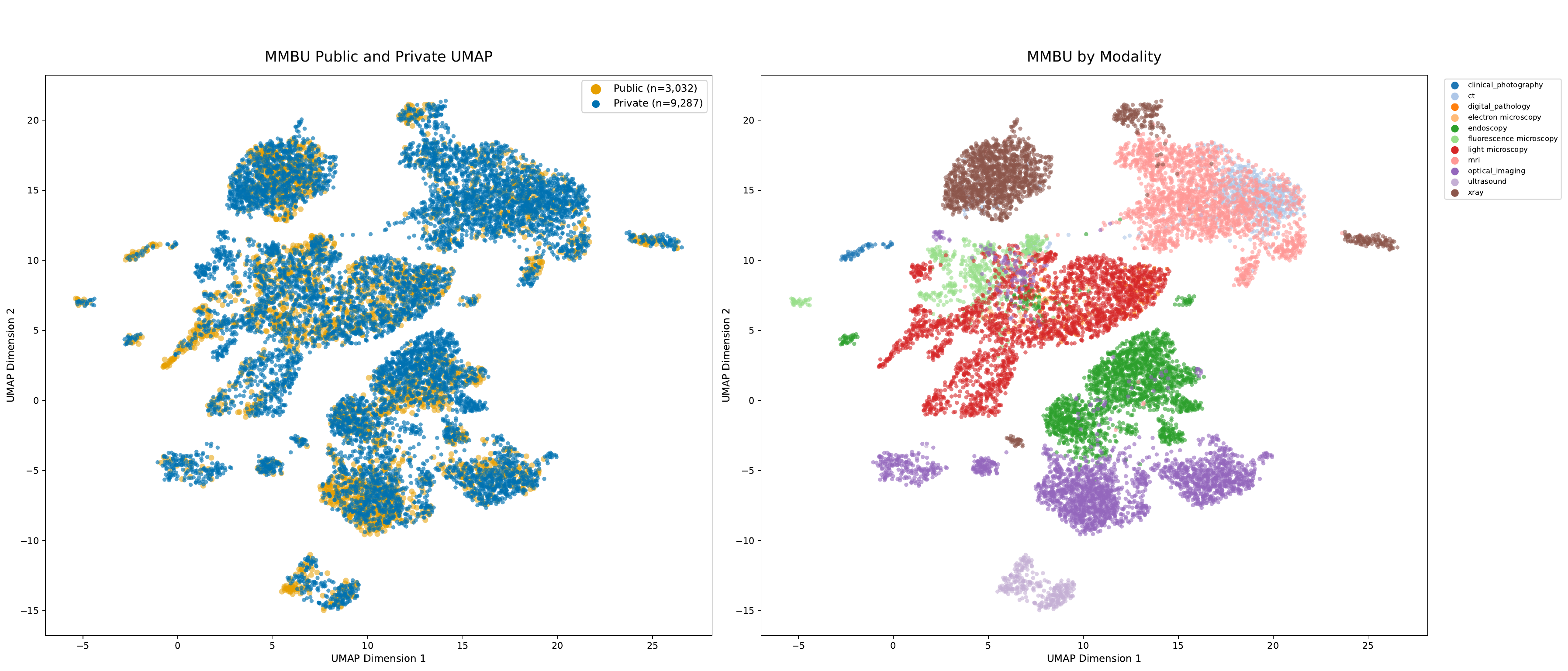}
    \caption{\textbf{Public/private split and modality coverage in MMBU.} Left: UMAP projection of DINOv3 embeddings, with the plotted public samples ($n=3{,}032$) shown in orange and private samples ($n=9{,}287$) in blue. Right: the embedding visualization colored by imaging modality. Counts refer to samples shown in this figure, rather than the total VQA item counts reported in the dataset description.}
    \label{fig:umap_public_private}
\end{figure*}

Figure~\ref{fig:umap_public_private} visualizes DINOv3 image embeddings for the public and private samples included in the plot. The left panel colors samples by split, while the right panel colors them by imaging modality. Public and private samples occupy many of the same regions in the displayed projection. This qualitative overlap illustrates shared visual coverage but does not establish that the two splits have identical distributions or difficulty.

\section{VLM Families for Evaluation}
\label{sec:supp_eval_vlms}
Table~\ref{tab:supp_eval_vlms} lists the 21 VLMs included in the main F1 comparison (Table~\ref{tab:model_comparison_total}). Table~\ref{tab:frontier_full_context} separately reports eight-task full-context results for eight models, including GPT-6-Sol and GPT-6-Luna, and public/private hard-set accuracies for nine models. GPT-6.1-Sol is included only in the hard-set comparison. Public/private accuracies for these three OpenAI models are from the project leaderboard~\cite{mmbu_leaderboard_2026}; none of the three is part of the 21-model F1 comparison. 

\begin{table*}[t]
\centering
\caption{\textbf{Evaluated VLMs on MMBU:} Architecture metadata for the 21 models in Table~\ref{tab:model_comparison_total}, grouped into medically adapted open-weight models, general-purpose open-weight models, and proprietary models. Dashes indicate unreported or inapplicable entries.}
\label{tab:supp_eval_vlms}
\begin{adjustbox}{width=\linewidth}
\begin{tabular}{lccccccc}
\toprule
\textbf{Model} & 
\textbf{Total Params} & 
\textbf{Base Framework} & 
\textbf{Vision Encoder} & 
\textbf{VE Params} & 
\textbf{Language Model} & 
\textbf{LM Params} & 
\textbf{Training Data} \\
\midrule
\rowcolor{mmbu!10} \multicolumn{8}{c}{\textbf{Open-Weight Medical Autoregressive VLMs}} \\

OctoMed-7B~\cite{ossowski2025octomed} & 7B & Qwen2.5-VL & Qwen2.5 ViT & - & Qwen2.5 LLM & 7B & Multiple \\
OctoMed-2-8B~\cite{ossowski2025octomed} & 8B & Qwen3-VL & SigLIP2-SO-400M & 400M & Qwen3 LLM & 8B & Multiple \\
Lingshu-7B~\cite{xu2025lingshu} & 7B & Qwen2.5-VL & Qwen2.5 ViT & - & Qwen2.5 LLM & 7B & Multiple \\
Lingshu-32B~\cite{xu2025lingshu} & 32B & Qwen2.5-VL & Qwen2.5 ViT & - & Qwen2.5 LLM & 32B & Multiple \\
LLaVA-Med-7B~\cite{li2024llava} & 7B & LLaVA & CLIP ViT-L/14 & 428M & Vicuna & 7B & PMC-15M \\
MedGemma-4B~\cite{sellergren2025medgemma} & 4B & Gemma 3 & MedSigLIP & 400M & Gemma3 LLM & 4B & Multiple \\
MedGemma-1.5-4B~\cite{sellergren2025medgemma} & 4B & Gemma 3 & MedSigLIP & 400M & Gemma3 LLM & 4B & Multiple \\

\rowcolor{mmbu!20} \multicolumn{8}{c}{\textbf{Open-Weight General-Purpose Autoregressive VLMs}} \\
Gemma-3-4B~\cite{gemma3technicalreport} & 4B & - & SigLIP & 400M &  Gemma3 LLM & 4B & Multiple \\

InternVL3.5-8B~\cite{internvl35} & 8B & - & InternViT & 300M &  Qwen3 LLM & 8B & Multiple \\

LLaVA-v1.5-7B~\cite{li2024llava} & 7B & - & OpenCLIP ViT-L/14 & 300M &  Vicuna-1.5 & 7B & Multiple \\
Qwen2.5-VL-3B-Instruct~\cite{qwen25vltechnicalreport} & 3B & - & Qwen2.5 ViT & - & Qwen2.5 LLM & 3B & Multiple \\
Qwen2.5-VL-7B-Instruct~\cite{qwen25vltechnicalreport} & 7B & - & Qwen2.5 ViT & - & Qwen2.5 LLM & 7B & Multiple \\
Qwen2.5-VL-32B-Instruct~\cite{qwen25vltechnicalreport} & 32B & - & Qwen2.5 ViT & - & Qwen2.5 LLM & 32B & Multiple \\

Qwen3-VL-4B-Instruct~\cite{qwen3vltechnicalreport} & 4B & - & SigLIP2-Large & 300M & Qwen3 LLM & 4B & Multiple \\
Qwen3-VL-8B-Instruct~\cite{qwen3vltechnicalreport} & 8B & - & SigLIP2-SO-400M & 400M & Qwen3 LLM & 8B & Multiple \\
Qwen3-VL-32B-Instruct~\cite{qwen3vltechnicalreport} & 32B & - & SigLIP2-SO-400M & 400M & Qwen3 LLM & 32B & Multiple \\

\rowcolor{mmbu!30} \multicolumn{8}{c}{\textbf{Closed-Source Frontier Autoregressive VLMs}} \\
Claude Opus 5 \cite{anthropic_opus5_2026} & - & Claude & - & - & - & - & Proprietary \\
Claude Sonnet 5 \cite{anthropic_sonnet5_2026} & - & Claude & - & - & - & - & Proprietary \\
Claude Haiku 4.5 \cite{anthropic_haiku45_2025} & - & Claude & - & - & - & - & Proprietary \\
GPT-5.4-mini \cite{openai_gpt54mini_2026} & - & GPT & - & - & - & - & Proprietary \\
GPT-4.1-mini \cite{openai_gpt41mini_2025} & - & GPT & - & - & - & - & Proprietary \\

\bottomrule
\end{tabular}
\end{adjustbox}
\end{table*}

\section{Additional Evaluation}
\label{sec:supp_evaluation}

\subsection{Evaluation Protocols and Scope}
\label{sec:supp_protocols}
The main comparison (Table~\ref{tab:model_comparison_total}) reports task-level micro-F1 for 21 models, using deterministic closed-answer extraction and Qwen3-32B semantic judging for open textual answers. The image ablation (Table~\ref{tab:image_no_image_main}) reports legacy-protocol averages for five open-weight models with and without the image. The full-context comparison (Table~\ref{tab:frontier_full_context}) evaluates eight models, including GPT-6-Sol and GPT-6-Luna, using the same structured reasoning prompts, full metadata context, and Sonnet-5 judging, and reports an unweighted mean of eight task accuracies. The public and private hard-set scores refer only to open-ended items. We additionally include leaderboard public/private accuracies for GPT-6-Sol, GPT-6-Luna, and GPT-6.1-Sol, giving nine models in the hard-set comparison. These hard-set accuracies are single-pass, tool-free leaderboard results~\cite{mmbu_leaderboard_2026}. The eight-task results for GPT-6-Sol and GPT-6-Luna instead come from our full-context evaluation under the same protocol as the other six models. Only GPT-6.1-Sol lacks full-context and per-task scores in this comparison. Confidence intervals are not reported for the full-context or hard-set accuracies. These summaries describe distinct evaluations and should not be treated as measurements under one shared protocol.

\subsection{Performance by Task, Modality, and Submodality}

\begin{figure*}[h]
    \centering
    \includegraphics[width=\linewidth, trim=4em 0 0 0]{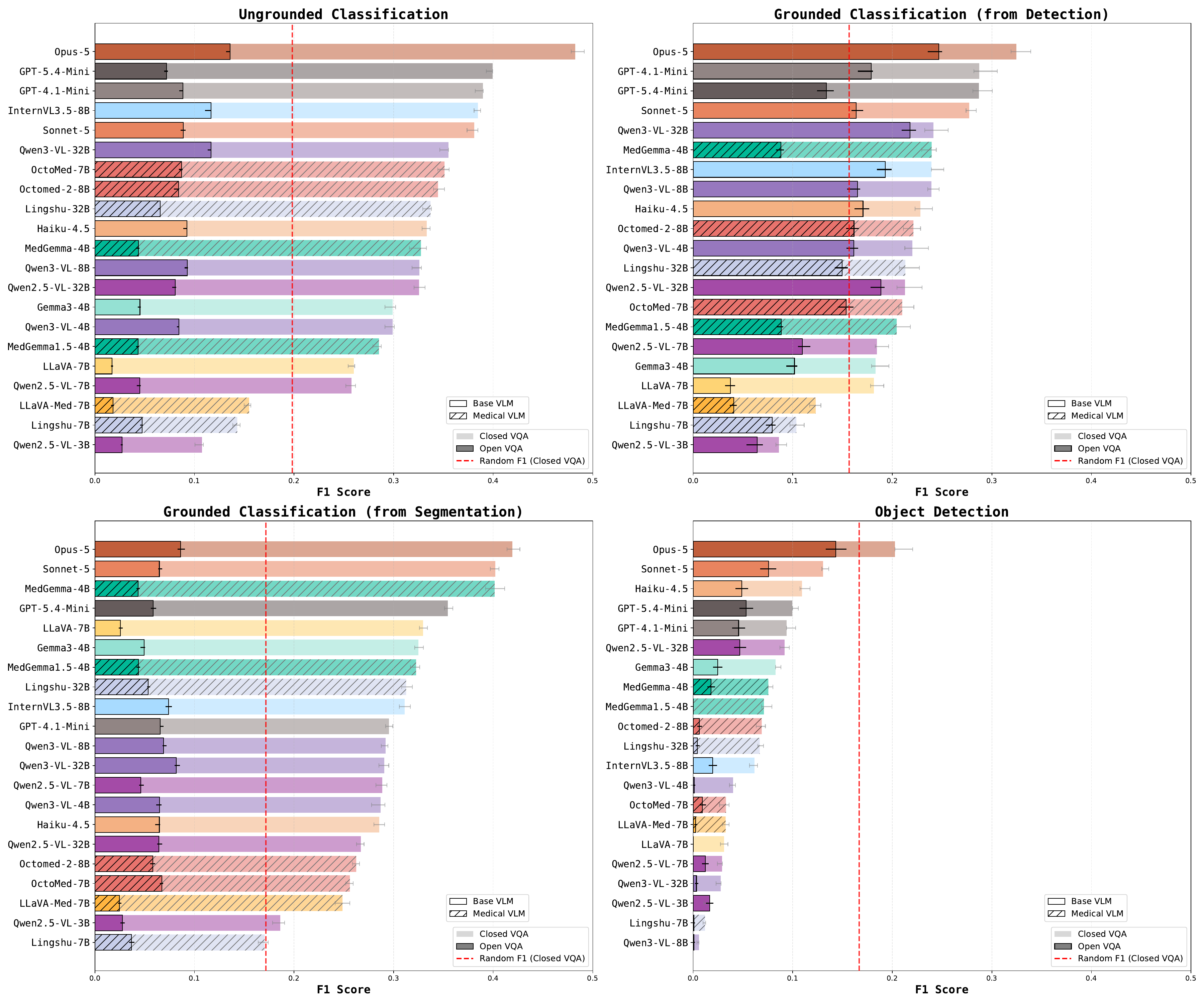}
\caption{\textbf{Aggregate performance on MMBU.} Performance of a representative set of VLMs on the \textit{classification} and \textit{detection} tasks in the benchmark. Solid outlines denote open-format results, while boxes without outlines denote closed-format results. Models are ranked by their closed-format performance. Dashes indicate models adapted to medical data. Similar colors indicate models of the same family.}
\label{fig:performance_by_task}
\end{figure*}

\begin{figure*}[h]
    \centering
    \includegraphics[width=\linewidth, trim=4em 0 0 0]{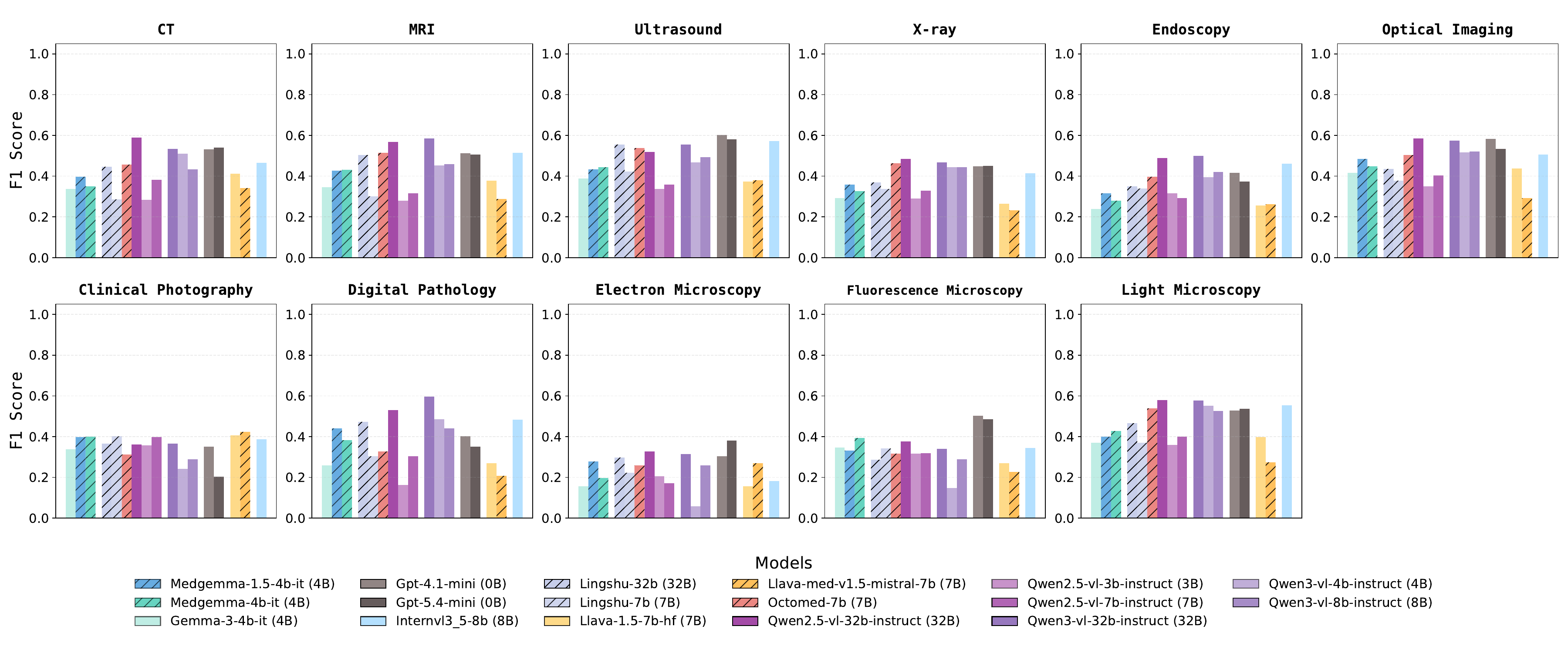}
    \caption{\textbf{Performance across 11 imaging modalities.} Each panel shows modality-specific F1 scores for the 17 models listed in the legend. Hatched bars identify medically adapted models; colors identify model families.}
    \label{fig:f1_modality}
\end{figure*}

Figure~\ref{fig:performance_by_task} separates classification and detection results by answer format. Figure~\ref{fig:f1_modality} further breaks performance down across the 11 imaging modalities. In the displayed results, scores are generally lower for electron microscopy than for MRI or light microscopy, and medical models do not improve uniformly over their comparison models. These descriptive differences motivate modality-specific reporting; they do not by themselves identify training-data coverage or quality as their cause.

\begin{figure*}[h]
    \centering
    \includegraphics[width=\linewidth, trim=4em 0 0 0]{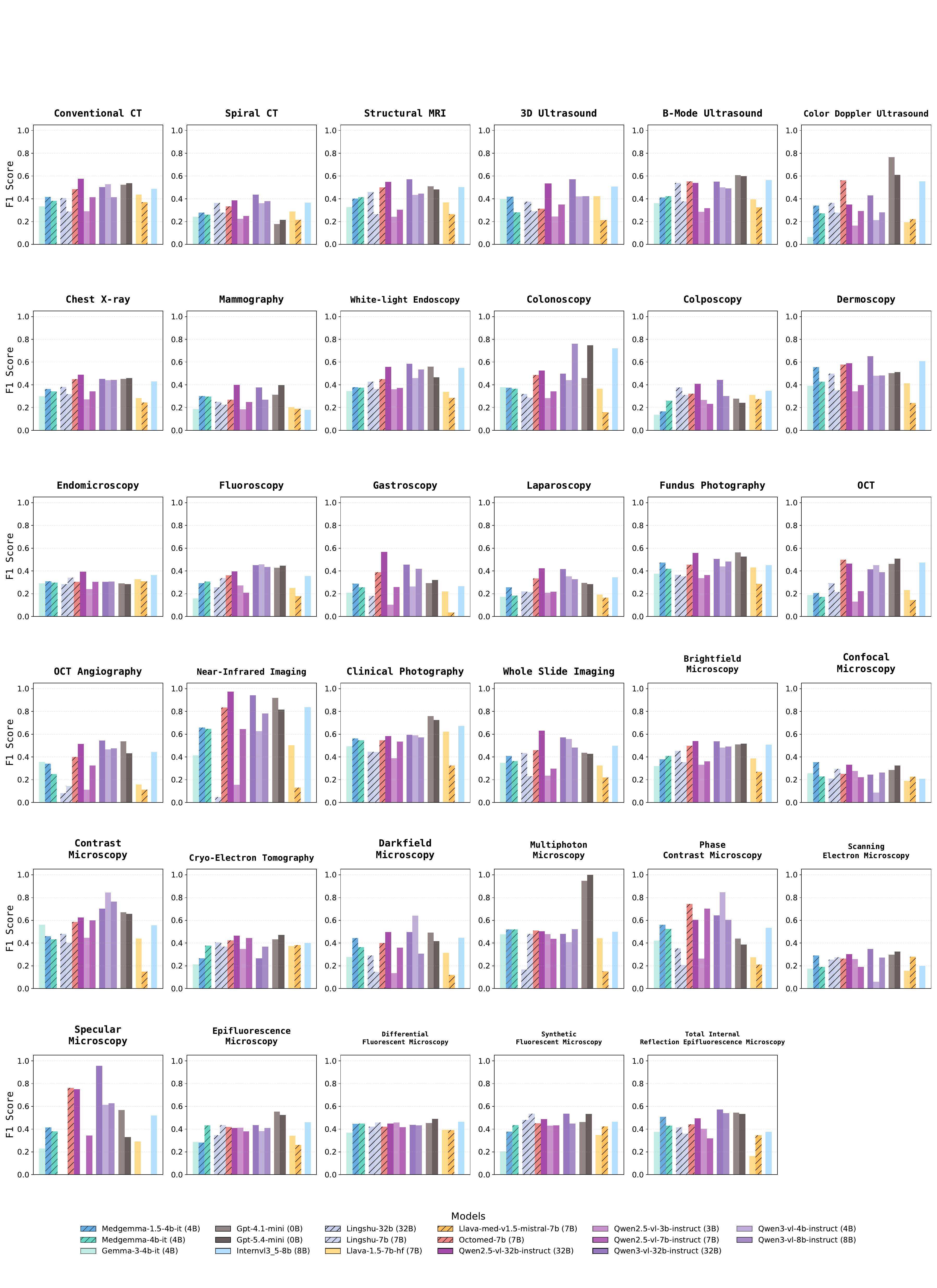}
    \caption{\textbf{Performance across 35 imaging submodalities.} Each panel shows submodality-specific F1 scores for the models in the legend. Colors and hatching follow Figure~\ref{fig:f1_modality}. The panels reveal variation within the broader modalities; category frequencies are needed to assess whether this variation is associated with sample size.}
    \label{fig:f1_submodality}
\end{figure*}

Figure~\ref{fig:f1_submodality} resolves differences within the broader modality groups. For example, the displayed scores are generally higher for near-infrared imaging than for OCT angiography, and the ranking of model families changes across microscopy submodalities. These comparisons identify settings in which performance differs, but should not be interpreted as evidence of a frequency-dependent long-tail effect without accounting for category sample sizes and uncertainty.

\subsection{Closed- and Open-Ended Answer Formats}
Table~\ref{tab:delta_f1_scores} summarizes equal-weight averages of the closed-minus-open differences displayed for the 21 models in Table~\ref{tab:model_comparison_total}. The largest mean gaps occur in segmentation-derived grounded classification (approximately 0.245) and ungrounded classification (0.241), followed by detection-derived grounded classification (0.077) and object detection (0.040). Closed scores are generally higher, although this pattern does not hold for every individual model-task pair. These differences capture the combined effects of answer-space constraints, output requirements, and scoring; they do not isolate visual grounding.


\definecolor{lightgray}{gray}{0.9}

\begin{table}[t]
    \centering
    \caption{Mean closed-minus-open F1 difference across the 21 models in Table~\ref{tab:model_comparison_total}, excluding the random baseline. Each model has equal weight. Values are calculated from the displayed, rounded delta entries and reported to three decimal places.}
    \label{tab:delta_f1_scores}
    \begin{tabular}{lc}
        \toprule
        \textbf{Task} & \textbf{Average $\Delta$ F1 Score} \\
        \midrule
        Ungrounded Classification & 0.241 \\
        Grounded Classification (from Detection) & 0.077 \\
        Object Detection & 0.040 \\
        Grounded Classification (from Segmentation) & 0.245 \\
        \bottomrule
    \end{tabular}
\end{table}

\begin{figure*}[t]
    \centering
    \includegraphics[
        width=\textwidth,
        height=0.82\textheight,
        keepaspectratio,
        trim=4em 0 0 0
    ]{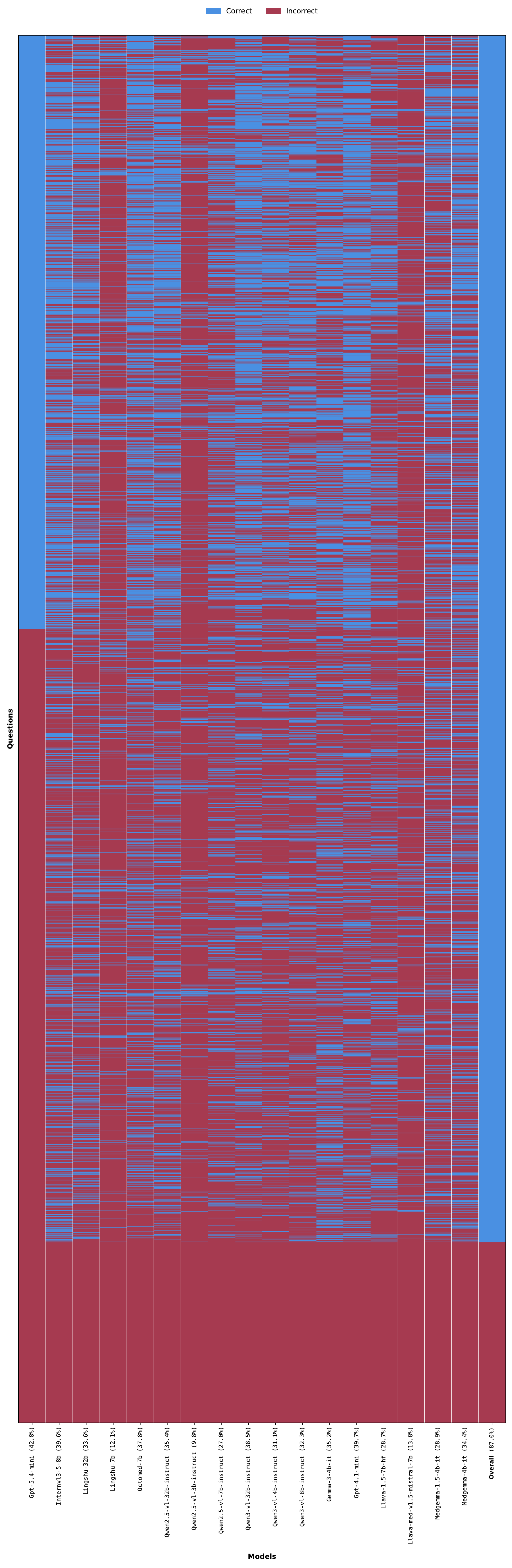}
    \caption{\textbf{Item-level correctness for the 17 displayed models.} Rows represent questions and model columns indicate correct (blue) or incorrect (red) responses. Percentages appear in the column labels.
    }
    \label{fig:correctness_heatmap}
\end{figure*}

\subsection{Domain-Level Comparisons of Medical and General-Purpose Models}
\begin{figure*}[t]
    \centering
    \includegraphics[width=\linewidth, page=1, trim=4em 0 0 0]{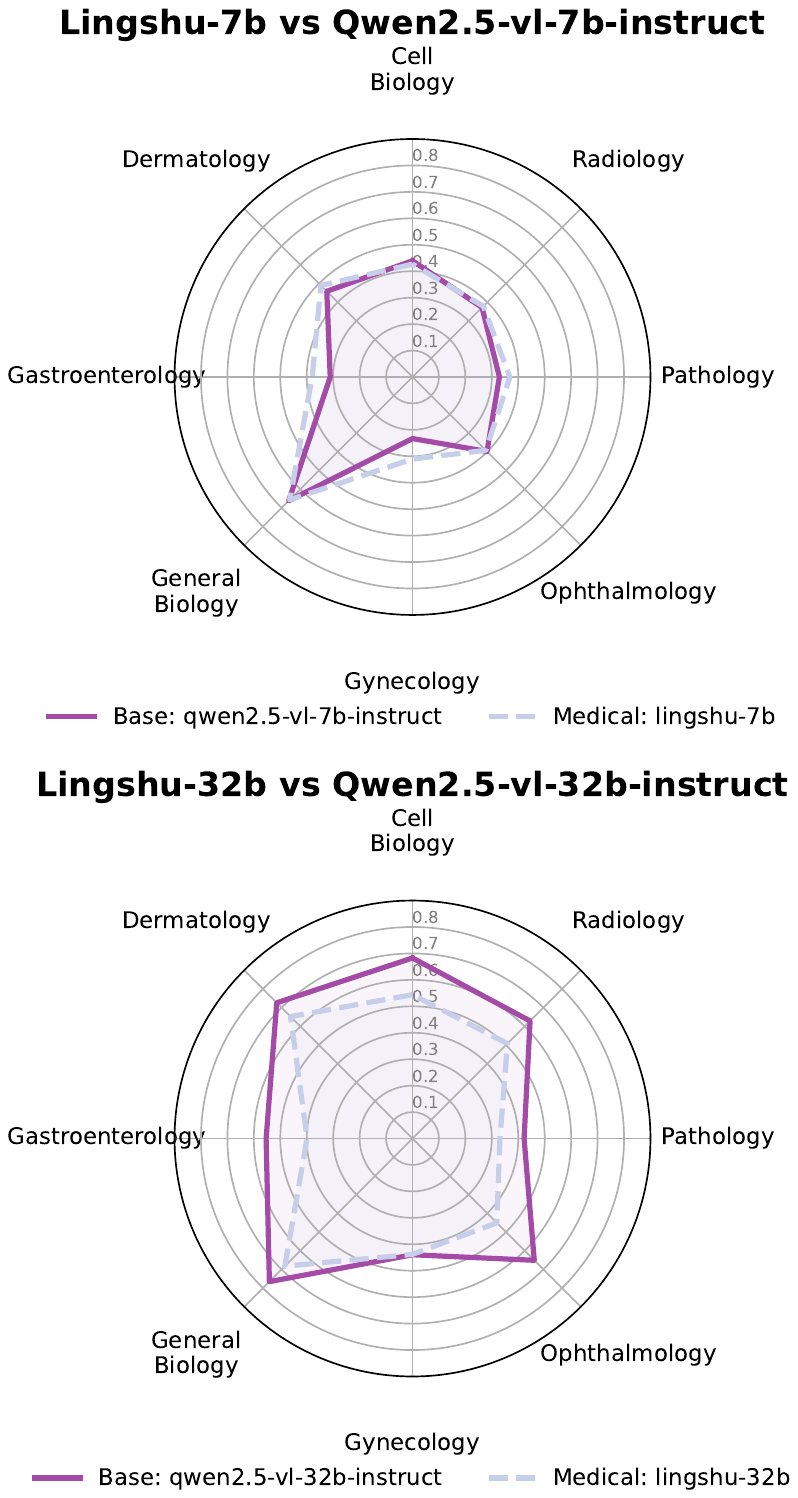}
    \caption{\textbf{Domain profiles for medical/general-purpose model pairs.} Each spoke denotes one of the eight displayed domains. Solid curves denote the general-purpose model and dashed curves the medical model; model names are given above each panel.}
    \label{fig:radial_p1}
\end{figure*}

Figures~\ref{fig:radial_p1}--\ref{fig:radial_p3} compare six medical/general-purpose model pairs across eight displayed biomedical domains. Spokes represent domains, solid lines represent general-purpose models, and dashed lines represent medically adapted models. The plots show that adaptation benefits vary by domain and model pair: the MedGemma and OctoMed comparisons exhibit improvements in several domains, while Lingshu-32B is below its comparison model in most domains.

Table~\ref{tab:model_comparison} summarizes the reported radial plot areas and their differences. This geometric summary is not an accuracy measure: it depends on axis scaling and domain ordering and need not lie between zero and one. We use it only as a descriptive comparison of the plotted profiles.

\definecolor{lightgray}{gray}{0.9}
\begin{table*}[h]
\caption{\textbf{Descriptive radar-area comparison.} Reported polygon areas for medical and general-purpose model pairs, with the medical-minus-general-purpose difference in the final column. Area is a geometric summary, not an accuracy or F1 score, and depends on domain ordering and axis scaling.}
\label{tab:model_comparison}
\centering
\renewcommand{\arraystretch}{1.2}
\begin{tabular}{l l c >{\columncolor{lightgray}}c}
\toprule
\textbf{Comparison Pair} & \textbf{Model Type} & \textbf{Area} & \textbf{Area Difference} \\
\midrule

\multirow{2}{*}{LLaVA-Med-v1.5} & Medical (LLaVA-Med) & 0.27 & \\
 & Base (LLaVA-v1.5) & 0.43 & \multirow{-2}{*}{-0.16} \\
\cmidrule{1-4}

\multirow{2}{*}{Lingshu-7b} & Medical (Lingshu-7b) & 0.50 & \\ 
 & Base (Qwen2.5-VL-7b) & 0.43 & \multirow{-2}{*}{+0.07} \\ 
\cmidrule{1-4}

\multirow{2}{*}{Lingshu-32b} & Medical (Lingshu-32b) & 0.70 & \\
 & Base (Qwen2.5-VL-32b) & 1.03 & \multirow{-2}{*}{-0.33} \\
\cmidrule{1-4}

\multirow{2}{*}{MedGemma-4b-it} & Medical (MedGemma) & 0.52 & \\
 & Base (Gemma-3-4b) & 0.35 & \multirow{-2}{*}{+0.17} \\
\cmidrule{1-4}

\multirow{2}{*}{MedGemma-1.5-4b-it} & Medical (MedGemma-1.5) & 0.55 & \\
 & Base (Gemma-3-4b) & 0.35 & \multirow{-2}{*}{+0.20} \\
\cmidrule{1-4}

\multirow{2}{*}{OctoMed-7b} & Medical (OctoMed-7b) & 0.83 & \\
 & Base (Qwen2.5-VL-7b) & 0.43 & \multirow{-2}{*}{+0.40} \\
\bottomrule
\end{tabular}
\end{table*}

\begin{figure*}[t]
    \centering
    \includegraphics[width=\linewidth, page=2, trim=4em 0 0 0]{combined_radial_plots.pdf}
    \caption{\textbf{Domain profiles for medical/general-purpose model pairs.} Each spoke denotes one of the eight displayed domains. Solid curves denote the general-purpose model and dashed curves the medical model; model names are given above each panel.}
    \label{fig:radial_p2}
\end{figure*}

\begin{figure*}[t]
    \centering
    \includegraphics[width=\linewidth, page=3, trim=4em 0 0 0]{combined_radial_plots.pdf}
    \caption{\textbf{Domain profiles for medical/general-purpose model pairs.} Each spoke denotes one of the eight displayed domains. Solid curves denote the general-purpose model and dashed curves the medical model; model names are given above each panel.}
    \label{fig:radial_p3}
\end{figure*}

\subsection{Effect of Question Context}
\label{sec:supp_context}
Figures~\ref{fig:radial_context_p1} and~\ref{fig:radial_context_p2} compare no-context, modality-only, and full-context prompts for GPT-5.4-mini and InternVL3.5-8B. Spokes represent the four task families, while colored curves represent context conditions; closed- and open-ended results appear in separate panels. For GPT-5.4-mini, full context provides a more visible improvement in open grounded classification than in closed classification. For InternVL3.5-8B, full context also improves the displayed closed detection score. The differing patterns across tasks indicate that additional metadata does not benefit every output format equally.

\begin{figure*}[t]
    \centering
    \includegraphics[width=\linewidth, page=3, trim=4em 0 0 0]{combined_radial_plots_all.pdf}
    \caption{\textbf{Effect of question context for GPT-5.4-mini.} Spokes denote task families; green, blue, and red curves denote no context, modality-only context, and full context, respectively. Closed-ended results are shown on the left and open-ended results on the right.}
    \label{fig:radial_context_p1}
\end{figure*}
\begin{figure*}[t]
    \centering
    \includegraphics[width=\linewidth, page=5, trim=4em 0 0 0]{combined_radial_plots_all.pdf}
    \caption{\textbf{Effect of question context for InternVL3.5-8B.} Task axes, context colors, and panel order follow Figure~\ref{fig:radial_context_p1}.}
    \label{fig:radial_context_p2}
\end{figure*}

\subsection{Image Ablation and Text-Only Baselines}
Table~\ref{tab:image_no_image_main} reports the paired image-conditioned and text-only results. Table~\ref{tab:closed_no_image_vs_random} compares the closed text-only scores against the reported random baseline. Three of the five models have point estimates above this baseline, but the mean difference across models is only 0.002. These estimates describe residual performance without images; uncertainty is needed to assess the reliability of individual differences.

\begin{table}[t]
\centering
\scriptsize
\setlength{\tabcolsep}{5pt}
\caption{Comparison between no-image closed VQA F1 scores and random performance. 
Random performance is 0.18. 
$\Delta$ is computed as F1$_{\text{no image}}$ - F1$_{\text{random}}$, where positive values indicate performance above random without image context.}
\label{tab:closed_no_image_vs_random}
\resizebox{\linewidth}{!}{
\begin{tabular}{lcc>{\columncolor{gray!15}}c}
\toprule
\textbf{Model} & \textbf{No Image Closed VQA} & \textbf{Random} & \textbf{$\Delta$} \\
\midrule

\textbf{OctoMed-7B} 
& 0.161 & 0.180 & -0.019 \\

\textbf{Qwen3-VL-32B} 
& 0.211 & 0.180 & +0.031 \\

\textbf{MedGemma-4B} 
& 0.110 & 0.180 & -0.070 \\

\textbf{MedGemma-1.5-4B} 
& 0.201 & 0.180 & +0.021 \\

\textbf{InternVL3.5-8B} 
& 0.226 & 0.180 & +0.046 \\

\midrule
\textbf{Average Across Models}
& 0.182 & 0.180 & +0.002 \\

\bottomrule
\end{tabular}
}
\end{table}

\subsection{Embedding Visualizations}
\begin{figure*}[t]
    \centering
    \includegraphics[width=\linewidth, page=1, trim=4em 0 0 0]{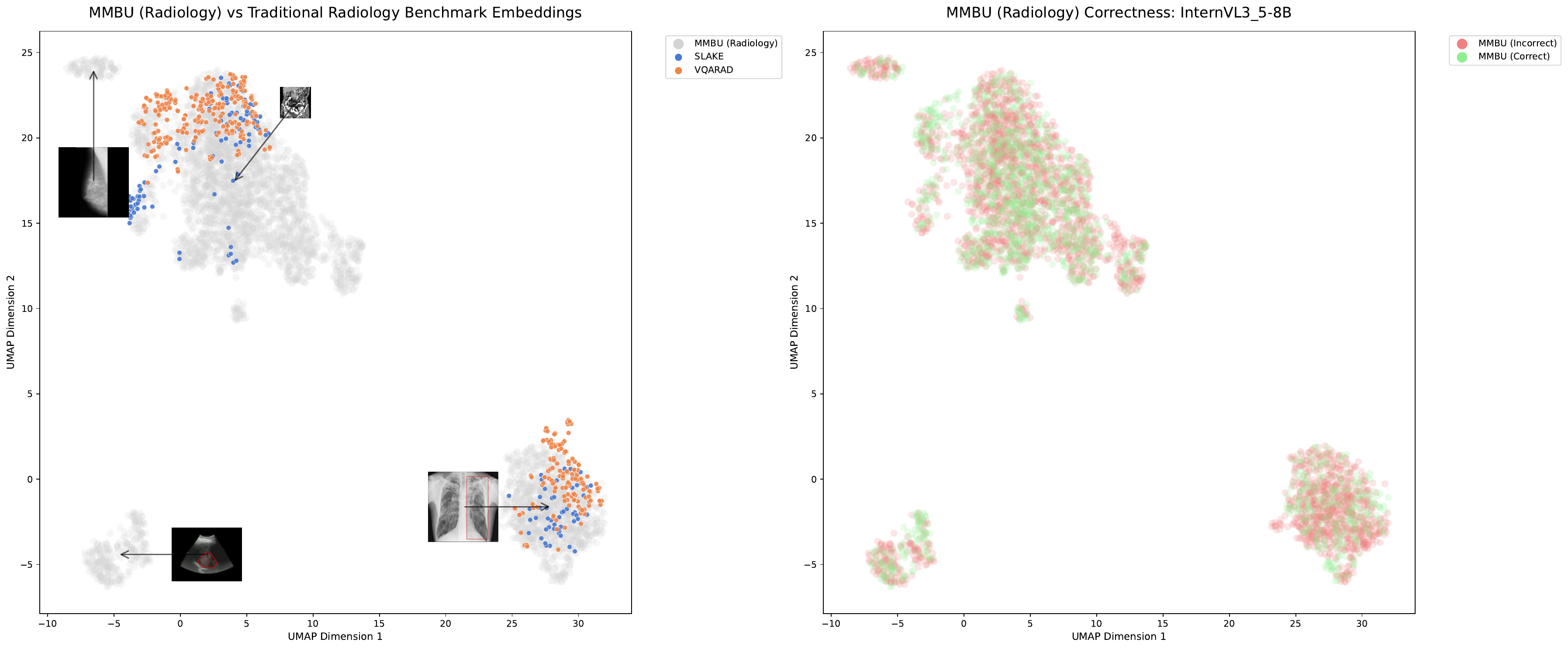}
    \caption{\textbf{Radiology embedding visualization.} Left: MMBU radiology examples alongside SLAKE and VQA-RAD examples. Right: MMBU examples colored by correctness for InternVL3.5-8B. The projection is a qualitative comparison of the displayed embeddings.}
    \label{fig:umap_rad_p1}
\end{figure*}

Figures~\ref{fig:umap_rad_p1}--\ref{fig:umap_path_p2} visualize MMBU radiology and pathology embeddings alongside established benchmarks. The left panels compare benchmark membership, and the right panels show correctness for the indicated model. Radiology includes both visually overlapping regions and regions occupied predominantly by MMBU examples in the plotted representation. These two-dimensional projections provide a qualitative view of coverage and errors. Visual overlap or separation alone does not establish membership in the same underlying distribution or prove out-of-distribution generalization.

\begin{figure*}[t]
    \centering
    \includegraphics[width=\linewidth, page=2, trim=4em 0 0 0]{rad_umap_combined.pdf}
    \caption{\textbf{Radiology embedding visualization.} Left: MMBU radiology examples alongside SLAKE and VQA-RAD examples. Right: MMBU examples colored by correctness for MedGemma-4B-it. The projection is a qualitative comparison of the displayed embeddings.}
    \label{fig:umap_rad_p2}
\end{figure*}

\begin{figure*}[t]
    \centering
    \includegraphics[width=\linewidth, page=1, trim=4em 0 0 0]{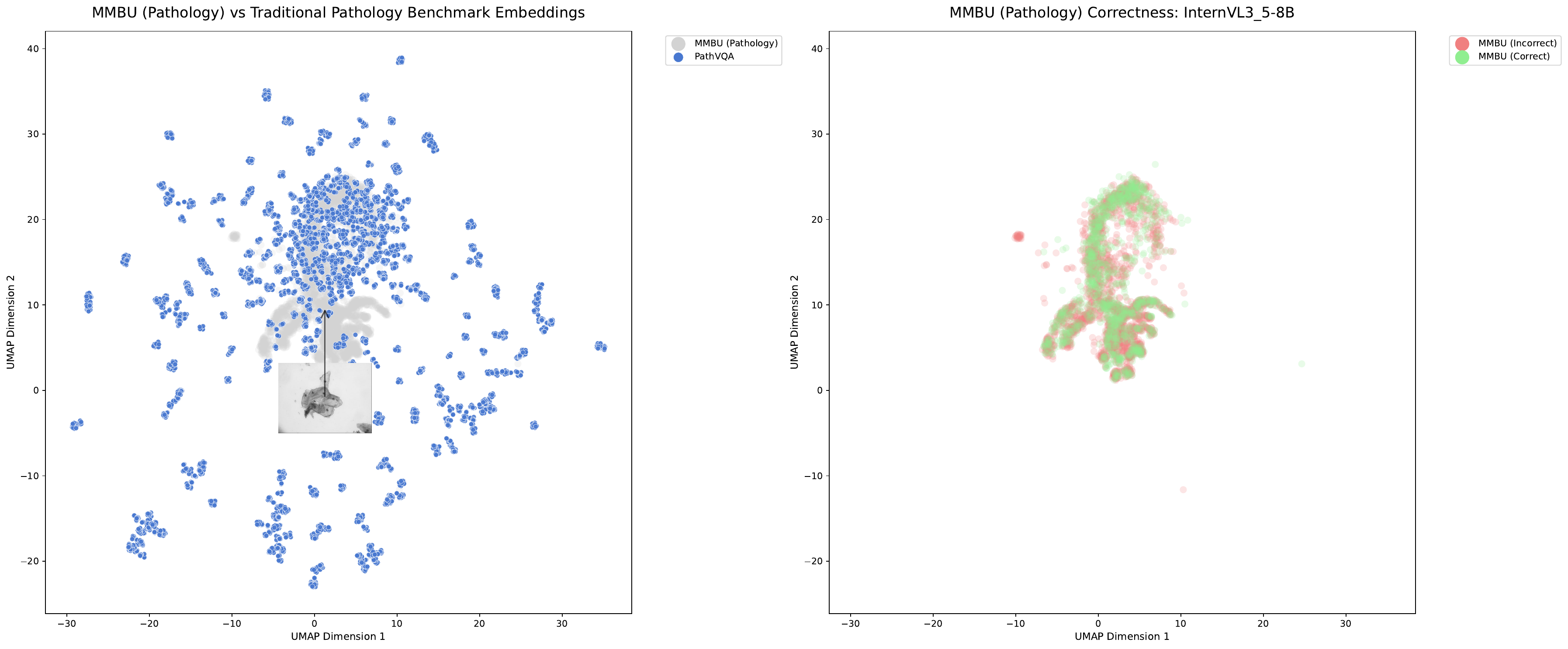}
    \caption{\textbf{Pathology embedding visualization.} Left: MMBU pathology examples alongside PathVQA examples. Right: MMBU examples colored by correctness for InternVL3.5-8B. Visual overlap in this projection does not establish equivalence of the underlying distributions.}
    \label{fig:umap_path_p1}
\end{figure*}
\begin{figure*}[t]
    \centering
    \includegraphics[width=\linewidth, page=2, trim=4em 0 0 0]{path_umap.pdf}
    \caption{\textbf{Pathology embedding visualization.} Left: MMBU pathology examples alongside PathVQA examples. Right: MMBU examples colored by correctness for MedGemma-4B-it. Visual overlap in this projection does not establish equivalence of the underlying distributions.}
    \label{fig:umap_path_p2}
\end{figure*}

\subsection{Matched-Image Comparison with GMAI-MMBench}
\begin{figure*}[t]
    \centering
    \includegraphics[width=\linewidth, page=1, trim=4em 0 0 0]{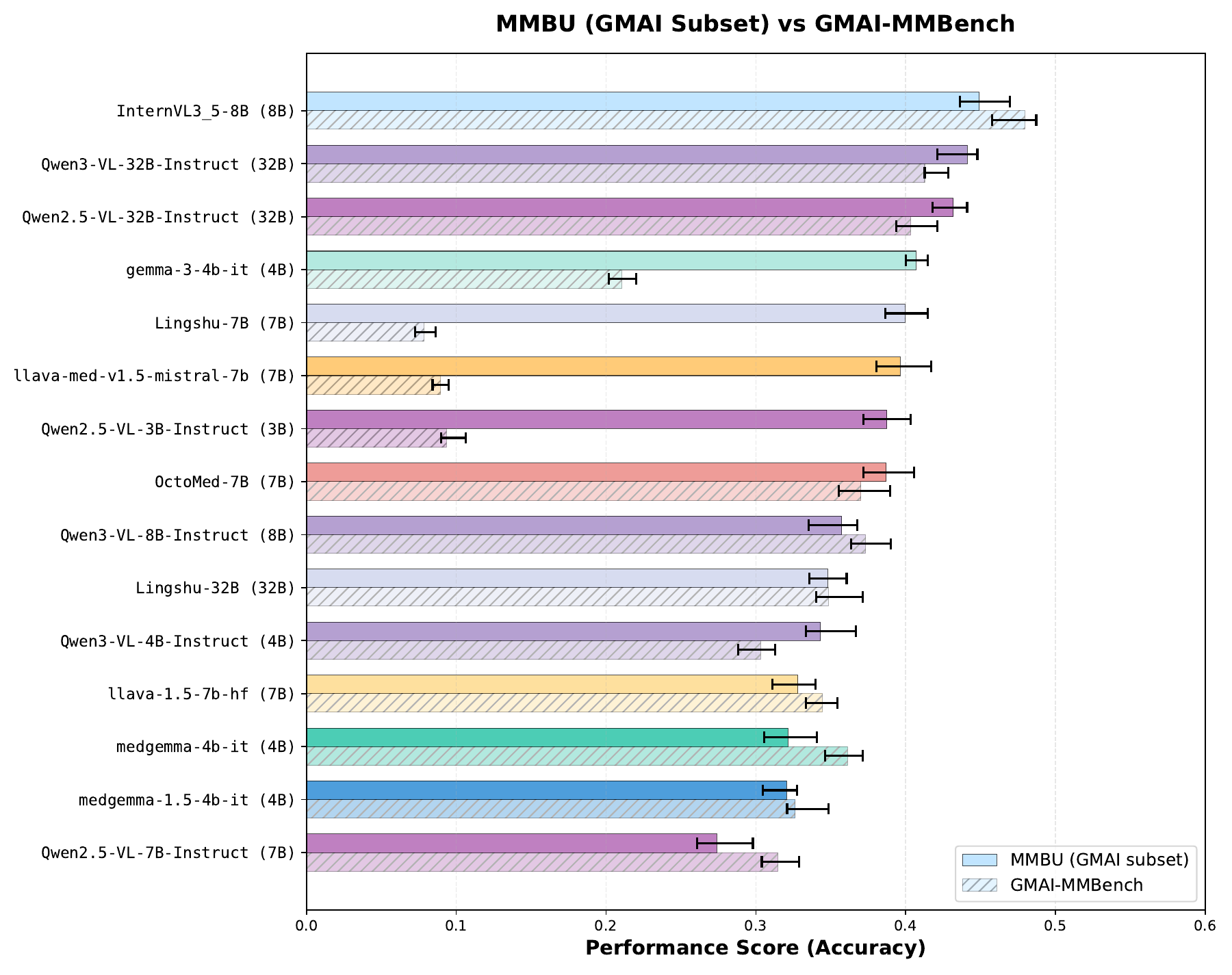}
    \caption{\textbf{Model Performance on GMAI-MMBench vs MMBU (GMAI Subset).} The accuracy of MMBU images from GMAI-MMBench with newly written full context is compared against GMAI-MMBench questions for the same subset of images.}
    \label{fig:gmai_vs_mmbu}
\end{figure*}

Figure~\ref{fig:gmai_vs_mmbu} compares responses to GMAI-MMBench questions with responses to newly written MMBU full-context questions for the same image subset. The plotted differences vary across models, with larger gains for some models than others. Because image identity is shared, this comparison examines the effect of question formulation and associated evaluation choices rather than isolating image-domain generalization.

\end{document}